\documentclass[acmsmall]{acmart}

\AtBeginDocument{%
  \providecommand\BibTeX{{%
    \normalfont B\kern-0.5em{\scshape i\kern-0.25em b}\kern-0.8em\TeX}}}

\usepackage{xcolor}
\usepackage{ifthen}
\usepackage{tikz}
\usetikzlibrary{patterns}

\usepackage{amsmath}
\usepackage{balance}
\usepackage[binary-units=true]{siunitx} 
\usepackage[caption=false]{subfig}
\usepackage{verbatim}
\usepackage{comment}
\usepackage[caption=false]{subfig}
\usepackage{xcolor,colortbl}
\definecolor{pairedOneLightBlue}{HTML}{a6cee3}
\definecolor{pairedTwoDarkBlue}{HTML}{1f78b4}
\definecolor{pairedThreeLightGreen}{HTML}{b2df8a}
\definecolor{pairedFourDarkGreen}{HTML}{33a02c}
\definecolor{pairedFiveLightRed}{HTML}{fb9a99}
\definecolor{pairedSixDarkRed}{HTML}{e31a1c}
\definecolor{butter1}{rgb}{0.988,0.914,0.310}
\definecolor{butter2}{rgb}{0.929,0.831,0.000}
\definecolor{butter3}{rgb}{0.769,0.627,0.000}
\definecolor{orange1}{rgb}{0.988,0.686,0.243}
\definecolor{orange2}{rgb}{0.961,0.475,0.000}
\definecolor{orange3}{rgb}{0.808,0.361,0.000}
\definecolor{chocolate1}{rgb}{0.914,0.725,0.431}
\definecolor{chocolate2}{rgb}{0.757,0.490,0.067}
\definecolor{chocolate3}{rgb}{0.561,0.349,0.008}
\definecolor{chameleon1}{rgb}{0.541,0.886,0.204}
\definecolor{chameleon2}{rgb}{0.451,0.824,0.086}
\definecolor{chameleon3}{rgb}{0.306,0.604,0.024}
\definecolor{skyblue1}{rgb}{0.447,0.624,0.812}
\definecolor{skyblue2}{rgb}{0.204,0.396,0.643}
\definecolor{skyblue3}{rgb}{0.125,0.290,0.529}
\definecolor{plum1}{rgb}{0.678,0.498,0.659}
\definecolor{plum2}{rgb}{0.459,0.314,0.482}
\definecolor{plum3}{rgb}{0.361,0.208,0.400}
\definecolor{scarletred1}{rgb}{0.937,0.161,0.161}
\definecolor{scarletred2}{rgb}{0.800,0.000,0.000}
\definecolor{scarletred3}{rgb}{0.643,0.000,0.000}
\definecolor{aluminium1}{rgb}{0.933,0.933,0.925}
\definecolor{aluminium2}{rgb}{0.827,0.843,0.812}
\definecolor{aluminium3}{rgb}{0.729,0.741,0.714}
\definecolor{aluminium4}{rgb}{0.533,0.541,0.522}
\definecolor{aluminium5}{rgb}{0.333,0.341,0.325}
\definecolor{aluminium6}{rgb}{0.180,0.204,0.212}

\definecolor{blind_safe_one_scheme_three_colors}{RGB}{102,194,165}
\definecolor{blind_safe_two_scheme_three_colors}{RGB}{252,141,98}
\definecolor{blind_safe_three_scheme_three_colors}{RGB}{141,160,203}

\definecolor{blind_safe_one_scheme_four_colors}{RGB}{166,206,227}
\definecolor{blind_safe_two_scheme_four_colors}{RGB}{31,120,180}
\definecolor{blind_safe_three_scheme_four_colors}{RGB}{178,223,138}
\definecolor{blind_safe_four_scheme_four_colors}{RGB}{51,160,44}

\definecolor{blind_safe_one_scheme_five_colors}{RGB}{240,249,232}
\definecolor{blind_safe_two_scheme_five_colors}{RGB}{186,228,188}
\definecolor{blind_safe_three_scheme_five_colors}{RGB}{123,204,196}
\definecolor{blind_safe_four_scheme_five_colors}{RGB}{67,162,202}
\definecolor{blind_safe_five_scheme_five_colors}{RGB}{8,104,172}

\definecolor{blind_safe_one_scheme_seven_colors}{RGB}{118,42,131}
\definecolor{blind_safe_two_scheme_seven_colors}{RGB}{175,141,195}
\definecolor{blind_safe_three_scheme_seven_colors}{RGB}{231,212,232}
\definecolor{blind_safe_four_scheme_seven_colors}{RGB}{247,247,247}
\definecolor{blind_safe_five_scheme_seven_colors}{RGB}{217,240,211}
\definecolor{blind_safe_six_scheme_seven_colors}{RGB}{127,191,123}
\definecolor{blind_safe_seven_scheme_seven_colors}{RGB}{27,120,55}

\definecolor{yellow_one}{RGB}{255,255,212}
\definecolor{yellow_two}{RGB}{254,217,142}
\definecolor{yellow_three}{RGB}{254,153,41}
\definecolor{yellow_four}{RGB}{217,95,14}
\definecolor{yellow_five}{RGB}{153,52,4}

\definecolor{forestgreen}{RGB}{34, 139, 34}
\usepackage[flushleft]{threeparttable}
\usepackage{textcomp}

\usepackage[printonlyused, withpage, nohyperlinks]{acronym}
\usepackage{algorithm}
\usepackage{lipsum}
\usepackage{multicol}

\usepackage[noend]{algpseudocode}
\algnewcommand{\LeftComment}[1]{\(\triangleright\)#1} 
\algrenewcommand{\algorithmiccomment}[1]{{\color{forestgreen}\(\triangleright\)#1}}
\algnewcommand\algorithmicinput{\textbf{Input}}
\algnewcommand\algorithmicoutput{\textbf{Output}}
\algnewcommand\Input{\item[\algorithmicinput]}%
\algnewcommand\Output{\item[\algorithmicoutput]}%
\algdef{SE}[VARIABLES]{Variables}{EndVariables}
    {\algorithmicvariables}
    {\algorithmicend\ \algorithmicvariables}
\algnewcommand{\algorithmicvariables}{\textbf{Global variables}}

\usepackage{relsize}
\usepackage{amssymb}

\let\emptyset\varnothing

\usepackage{pgfplots,pgfplotstable}

\newcommand{\textoverline}[1]{$\overline{\mbox{#1}}$}

\setcopyright{acmcopyright}
\copyrightyear{2022}
\acmYear{2022}
\acmDOI{10.1145/1122445.1122456}

\acmJournal{TECS}
\acmVolume{XX}
\acmNumber{Y}
\acmArticle{1}
\acmMonth{6}

\pgfplotsset{compat=1.17} 
\begin{document}

\title[Brain-inspired Cognition in Next Generation Racetrack Memories]{Brain-inspired Cognition in Next Generation Racetrack Memories}

\author{Asif Ali Khan}
\authornote{Authors contributed equally to the paper.}
\email{asif_ali.khan@tu-dresden.de}
\orcid{0000-0002-5130-9855}
\affiliation{%
  \institution{Center for Advancing Electronics Dresden (cfaed), TU Dresden}
  \streetaddress{Helmholtzstrasse 18
}
  \postcode{01069}
  \city{Dresden}
  \country{Germany}}
\author{S{\'e}bastien Ollivier}
\authornotemark[1]
\email{sbo15@pitt.edu}
\orcid{XXXX}
\author{Stephen Longofono}
\email{stl77@pitt.edu}
\affiliation{%
  \institution{University of Pittsburgh}
  \streetaddress{1238 Benedum Hall, 3700 O'Hara Street}
  \city{Pittsburgh}
  \state{Pennsylvania}
  \country{USA}
  \postcode{15260}
}
\author{Gerald Hempel}
\email{gerald.hempel@tu-dresden.de}
\author{Jeronimo Castrillon}
\email{jeronimo.castrillon@tu-dresden.de}
\orcid{0000-0002-5007-445X}
\affiliation{%
  \institution{Center for Advancing Electronics Dresden (cfaed), TU Dresden}
  \streetaddress{Helmholtzstrasse 18}
  \postcode{01069}
  \city{Dresden}
  \country{Germany}}

\author{Alex K. Jones}
\email{akjones@pitt.edu}
\affiliation{%
  \institution{University of Pittsburgh}
  \streetaddress{1238 Benedum Hall, 3700 O'Hara Street}
  \city{Pittsburgh}
  \state{Pennsylvania}
  \country{USA}
  \postcode{15260}
}

\renewcommand{\shortauthors}{Khan and Ollivier et al.}

\begin{abstract}

\emph{Hyperdimensional computing} (HDC) is an emerging computational framework inspired by the brain that operates on vectors with thousands of dimensions to emulate cognition. Unlike conventional computational frameworks that operate on numbers, HDC, like the brain, uses high dimensional random vectors and is capable of one-shot learning. 
HDC is based on a well-defined set of arithmetic operations and is highly error-resilient. 
The core operations of HDC manipulate HD vectors in bulk bit-wise fashion, offering many opportunities to leverage parallelism. 
Unfortunately, on conventional von Neumann architectures, the continuous movement of HD vectors among the processor and the  memory can make the cognition task prohibitively slow and energy-intensive. Hardware accelerators only marginally improve related metrics. In contrast, even partial implementations of an HDC framework inside memory can provide considerable performance/energy gains as demonstrated in prior work using memristors. This paper presents an architecture based on \emph{racetrack memory} (RTM) to conduct and accelerate the entire HDC framework within memory. The proposed solution requires minimal additional CMOS circuitry by leveraging a read operation across multiple domains in RTMs called \emph{transverse read} (TR) to realize exclusive-or (\texttt{XOR}) and addition operations. To minimize the CMOS circuitry overhead, an RTM nanowire-based counting mechanism is proposed. Using language recognition as the example workload, the proposed RTM HDC system reduces the energy consumption by 8.6$\times$ compared to the state-of-the-art in-memory implementation.  Compared to dedicated hardware design realized with an FPGA, RTM-based HDC processing demonstrates 7.8$\times$ and 5.3$\times$ improvements in the overall runtime and energy consumption, respectively.
\end{abstract}



\maketitle

\section{Introduction}
\label{sec:intro}
The success of machine learning has fueled the transformation of industry and society in recent decades. A key factor for the ubiquity of these learning algorithms is their use in mobile devices such as smartphones, tablets, or sensor networks. However, classic approaches such as deep learning require enormous computing and power resources~\cite{ML_compute_resources}. For example, training of a single transformer-based deep learning model requires weeks on modern GPUs and produces carbon footprints (a proxy for energy consumption) $\approx 5\times$ more than the entire lifetime carbon footprint of a passenger car~\cite{NLP_models_energy}. 
Unfortunately, these characteristics are at odds with the requirements of many IoT devices, namely limited bandwidth, memory and compute power, and battery capacity. Architectural innovations such as near-memory and in-memory computing, along with the alternate models for machine learning such as hyperdimensional computing, substantially reduce the area and energy consumption of cognitive-inspired computing systems without compromising accuracy~\cite{inPCM_2020}.

The idea of \ac{HDC} is inspired by biological systems that generally combine sufficient accuracy with a very high energy efficiency. Compared to conventional machine learning models, HDC is more robust and error-resilient~\cite{kanerva_2009} as well as more compute and energy efficient~\cite{hdc_review, HDCFPGA}. Moreover, HDC provides comparable accuracy to the highest fidelity ML models (cf. Table 2 in \cite{hdc_comp}).
 HDC frameworks mainly operate on binary or bi-polar hypervectors, typically having thousands of dimensions~\cite{kanerva_2009}. The \emph{base} or \emph{seed} hypervectors are randomly generated and describe input features. In HDC training, class hypervectors are generated by performing a set of basic algebraic operations (XOR, permutation, addition, thresholding, and multiplication) that combine several hypervectors and the properties of the desired class. In inference, the same encoding is applied to the input data to generate a query hypervector and reason about a given dataset. The query hypervector is then classified by performing a similarity match operation.

With conventional von Neumann machines, shuttling of hypervectors between the memory and the processor makes the overall classification process prohibitively slow. To overcome this, state-of-the-art proposals use accelerators and near-memory processing to achieve parallelism and energy efficiency~\cite{rahimi_ISLPED_16, rahimi_TCS_17,  hdcASIC_2019}. Since the algebraic operations in most of the HDC frameworks are memory intensive and inherently parallel, they are particularly well-suited for in-memory computing. Furthermore, in most emerging memory technologies, the physical properties of the memory cells can be exploited to realize some, if not all, HDC operations in place~\cite{cim_resistive_2018, cim_rtm_2021}. 

In one of the most recent works, an entire HDC framework is implemented on an integrated system using memristor crossbars with additional CMOS logic~\cite{inPCM_2020}. Specifically, the multiplication operation required for ``binding'' and ``similarity search'' operations is implemented using phase change memory (PCM) crossbars while the addition, permutation and thresholding operations are realized by additional near-memory CMOS logic.  Although the in-PCM HDC system significantly reduces energy consumption (by more than $6\times$), it has three major limitations. First, the additional CMOS logic incurs large area and energy penalties. In the ideal case, the entire framework should be implemented using memory devices. Second, the write operation in resistive memories such as PCM is extremely expensive (in terms of latency and energy) and induces wear on the endurance-limited cells. Although the proposed solutions avoid repetitive programming of the memristive devices, the fundamental problem of expensive writes and finite endurance remains. Third, memristive devices compute values in the analog domain. Besides accuracy implications, which are not as severe due to the inherent resilience of HDC, analog computation requires power hungry~\cite{isaac_2016} back-and-forth conversion between the analog and digital domains (via ADC/DAC).

To overcome these challenges, we use another class of emerging nonvolatile memory technologies called \ac{RTM} or \ac{DWM}~\cite{blaesing_2020} to implement the entire HDC framework. An RTM cell consists of a magnetic nanowire that stores multiple data bits in magnetic \emph{domains} and is associated with one or more access ports. RTM promises to realize the entire framework in the digital domain with relatively low additional logic and
without compromising on accuracy. 

We present \ac{HDCR}, 
a complete in-RTM HDC system where all HDC operations are implemented in RTM using the RTM device characteristics. 
Namely, a novel access mode called \ac{TR}~\cite{roxy2020novel} is used to conduct processing within the RTM~\cite{ollivier2019dsn,CORUSCANT}. By applying a sub-shift-threshold current across two access points along the nanowire, the resistance state of the nanowire can be used to count `1's at each bit position across multiple adjacent data words within the memory.
HDCR leverages the TR operation and makes appropriate changes to the peripheral circuitry to realize the XOR operation, and efficient counters. Together with our design for in-memory 
majority operation, and ``permutation,'' TR enables all necessary HDC processing operations to be performed in a highly parallel fashion within RTM.

Our experimental results show that for the well-known use case of language recognition, our HDC system is an order of magnitude faster than the state-of-the-art FPGA solution and consumes 5.3$\times$ and $8.6\times$ less energy compared to the state-of-the-art FPGA and PCM-crossbar solutions, respectively.

~\\
The main contributions of this paper are as follows:

\begin{enumerate}
    \item We present a complete HDC system with precise control and datapaths based on nonvolatile racetrack memory.
    \item For the rotation operation, we make necessary changes to the RTM row buffer to enable rotation of HD vectors with a simple copy (read and write) operation.
    \item We propose a first RTM nanowires-based counter design to perform the majority operation and compute the Hamming weight. 
    \item For binding, we implement the XOR logic by doing a transverse read operation and using the modified row buffer to infer the result.  
    \item For bundling, we use RTM counters to find the majority output at each position in the hypervectors.
    \item For comparison with the class vectors, we compute the Hamming distance between the query vector and each class vector leveraging a TR-based XOR operation and the RTM counter. 
    \item We evaluate our system on a standard benchmark and compare the runtime and energy consumption with state-of-the-art FPGA~\cite{rahimi_TCS_17} and in-PCM implementations~\cite{inPCM_2020}. 
\end{enumerate}

The remainder of this paper is organized as follows: Section~\ref{sec:background} provides background information about HDC, language recognition, RTM and TR. Section~\ref{sec:cim} proposes the architectural modification needed to perform operations inside RTM and explains the implementation of our RTM counter. Section~\ref{sec:HDCInRacetrack} explains different HDCR modules and their  integration to perform HDC operations in RTM. Section~\ref{sec:Evaluation} evaluates HDCR, demonstrating the energy and latency advantages of using RTM. Section~\ref{sec:related_work} presents some of the most related work in the literature. Finally, Section~\ref{sec:Conclusion} concludes the paper.  

\section{Background}
\label{sec:background}
In this section, we introduce the fundamentals of HDC, its major operations, and main components. We then describe our use case and provide details on classes and input features/symbols.
Finally, we provide background on \ac{RTM} technology, its properties and organization, and the working principles of the transverse read operation. 

\subsection{Hyperdimensional Computing}
\label{subsec:hdc_bg}
Hyperdimensional computing, also referred to as brain-inspired computing, is based on the observation that neural activity patterns can be regarded as one fundamental component behind cognitive processes. These patterns can be modeled by leveraging the mathematical properties of hyperdimensional spaces. In conjunction with a well-defined algebra, they can be used to implement machine learning tasks with less computational effort than other approaches such as the support vector machine (SVM) algorithm~\cite{2018svmvshdc}. Since the dimension $D$ of the hyperdimensional space is on the order of $10^4$, this approach is extremely robust to variation and errors within its hypervectors. 

In HD computing, each \ac{HV} describes a unique point in space and encodes either a feature, a group of features, or a class in the given machine learning problem. As shown in Fig.~\ref{fig:overview_hdc}-I, the base or seed hypervectors describe input features, and are randomly generated. In HDC training, a set of algebraic operations---\textit{i.e.}, binding, bundling, permutation, and similarity check---are performed on the seed hypervectors and their intermediate results are used to generate \emph{class} hypervectors. Each class hypervector represents a class in the data set. In HDC inference, the same encoding is applied to the input data to generate a \emph{query} hypervector. The query hypervector is then classified into one of the classes by performing a similarity check operation.

Various HDC frameworks exist that implement HDC in different ways such as (1) using different types of hypervectors (bipolar, binary, integer, etc.), (2) using a different distribution of elements in hypervectors (sparse and dense hypervectors), and (3) employing a different set of algebraic operations. A detailed comparison of these frameworks is presented in~\cite{rahimi_TCS_17, vsas}. Since we focus on a digital, in-memory implementation of HDC, we consider a binary HDC subset. Thus hypervectors consist of binary values and the framework leverages Boolean operations to implement the required algebraic operations. For the hypervectors, we consider the dimensionality of a hypervector $D=8192$ and a probability of $P=0.5$ for each component to be a one or a zero. This is because, for our selected use case, $D=8192$ does not have any considerable impact on the accuracy (only reduces it from 97.8\% to 97.7\%) while still leaving the memory to be used by other general-purpose applications.

We use the Hamming distance  $d_H(\vec{a},\vec{b})$ metric to compare the hypervectors $\vec{a}$ and $\vec{b}$, resulting in the normalized number of dissimilar elements of both vectors. For large vector sizes, the Hamming distance between random vector pairs, in 98\% of the cases, results in $d_H(\vec{a},\vec{b}) = D/2$. In this context, we classify any two vectors as similar ($d_H<0.5$) or dissimilar ($d_H \geq 0.5$). Since $d_H(\vec{a},\vec{b}) \approx B(D,P=1/2)$ with $B$ representing the binomial distribution, random, \textit{i.e.}, unrelated, vectors are unlikely to deviate from $D/2$.  Thus, HDC defines sufficiently dissimilar (\textit{e.g., }$d_H \geq 0.5$) vectors to be \textit{orthogonal}\footnote{Mathematically, orthogonal vectors would have $d_H=1$, HDC relaxes this definition to $d_H\geq0.5$ because it is attempting to distinguish between \textit{similar} and \textit{dissimilar} vectors.  HDC redefines vectors with $d_H=1$ as \textit{diametrically opposed}.}.    



In the context of HDC for binary hypervectors, relevant algebraic operations are: 

\begin{itemize}
  \item \textbf{Binding} is used for combining related hypervectors.
  This operation is implemented as an element-wise \texttt{XOR} operation between $N$ hypervectors \textit{e.g.}, $\vec{c} = \vec{x_1} \oplus \vec{x_2} \dots \vec{x_N}$ binds $\vec{x_i} : i = 1,2,...,N$ together. 
  \item \textbf{Permutation} is used to generate a new  hypervector that is orthogonal to the original hypervector by performing a reversible operation. The permutation is a unary operation $\vec{x_p}=\rho(\vec{x})$ such that the resulting vector $\vec{x_p}$ is orthogonal to $\vec{x}$. 
  In the context of this work, we use piece-wise circular shifts 
  to perform this operation (see Section~\ref{sss:binding}).
    Rotating a hypervector $n$ times is expressed as $\vec{x_p}=\rho^n(\vec{x})$.
  \item \textbf{Bundling} is used to generate a hypervector representing a set of hypervectors. This operation is implemented by performing the vector sum and element-wise thresholding,  also referred to as the majority operation. For an even number of binary hypervectors, the tie is broken by a fixed random hypervector. The bundling operation generates a representative hypervector which is non-orthogonal to the operand hypervectors.
    \item \textbf{Similarity Check}: The similarity check operation compares the query hypervector to all class hypervectors to find the closest match. Different frameworks use a variety of similarity metrics. For this work, we use Hamming distance and compare the Hamming weights of the query and class hypervectors. The operation is implemented as an \texttt{XOR} followed by the population count operation (see Section~\ref{subsec:inference}.) 
\end{itemize}

\subsection{Use Case: Language Recognition}
\label{subsec:use_case}
\begin{figure}[bp]
\centering
\includegraphics[width=\textwidth]{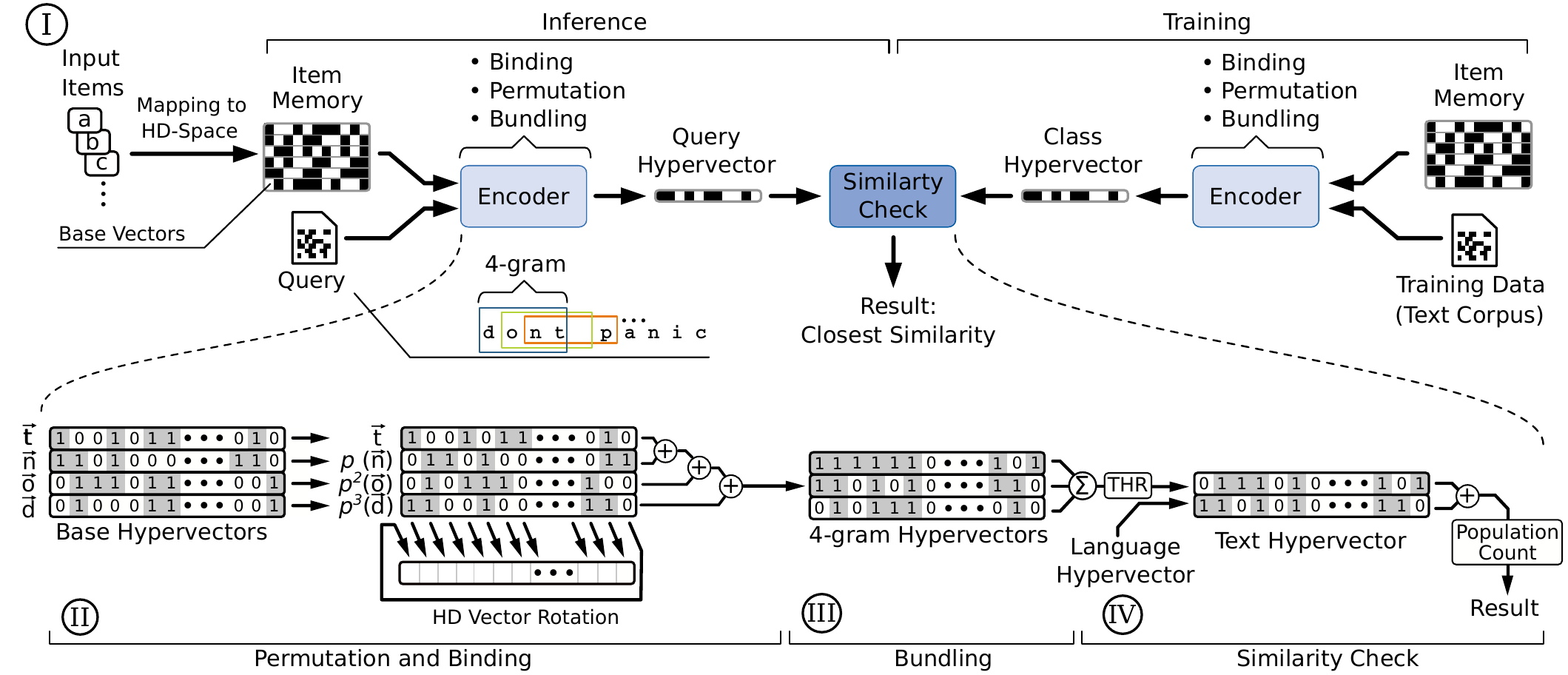}
\caption{An overview of the HDC operations}
\label{fig:overview_hdc}
\end{figure}

In the context of this work, we use the \ac{LR} classification task, which has already been used as a benchmark by other HDC approaches in the literature~\cite{rahimi_ISLPED_16, rahimi_TCS_17, inPCM_2020}. 
With this example application, we demonstrate the scalability and efficiency of our architecture compared to the state-of-the-art FPGA~\cite{rahimi_TCS_17} and in-memory~\cite{inPCM_2020} implementations. We use the language recognition code published on~\cite{git_sources} that classifies an input text to one of 22 European languages. The input features consist of 26 letters of the Latin alphabet and the space character (represented by $\tau$).  As a first step in building the hyperdimensional (HD) model, hypervectors are generated for all input letters and are stored in an \ac{IM} $\Theta = \{a \rightarrow \vec{a}, b \rightarrow \vec{b}, \dots, z \rightarrow \vec{z}, \tau \rightarrow \vec{\tau} \}$ (see Fig.~\ref{fig:overview_hdc}-I).  
The dimensionality of the hypervectors ($D=8192$) is carefully chosen to ensure better utilization of the memory architecture.


After the IM is created, the training of the HD model is carried out using one text for each of the 22 languages.  
In order to model the probability distribution of individual letters in the respective language, the text is broken down into substrings of length $N$ called \emph{N-grams}. In the binding operation, a hypervector is generated for each N-gram of the input text, which is subsequently combined by the bundling operation into a single hypervector. This is in contrast to models which use dictionaries and banks of phrases, which increases the complexity of similarity checking without a commensurate advantage in accuracy or efficiency~\cite{2010langmodel}. 
For example, the first N-gram of the phrase \emph{``dont panic''} for $N=4$ would be \emph{``dont''}. This is encoded to a single N-gram vector, as shown in Fig.~\ref{fig:overview_hdc}-II, by permuting and \texttt{XOR}ing the individual hypervectors from the IM ($\vec{\Theta}$) as follows: $\vec{\Phi}_{dont} = \rho^3(\vec{d}) \oplus \rho^2(\vec{o}) \oplus \rho(\vec{n}) \oplus \vec{t}$. 
Due to the properties of the selected encoding, all generated N-gram vectors $V_z = \{\vec{\Phi}_ {dont}, \vec{\Phi}_{ont\tau}, \dots, \vec{\Phi}_{anic} \}$ are orthogonal. 
Finally, the language vector $\vec{\mathcal{T}}$ is generated as follows: $\vec{\vec{\mathcal{T}}} = \text{Majority }( \vec{\Phi}_ {dont}, \vec{\Phi}_{ont\tau}, \dots, \vec{\Phi}_{anic})$ (see Fig.~\ref{fig:overview_hdc}-III).
In the training phase, $\vec{\mathcal{T}}$ represents a (language) class hypervector $\vec{\mathcal{L}}$ and is stored in the associative memory. In the inference phase of HDC, $\vec{\mathcal{T}}$, the query hypervector, represents the input sentences or phrases and is generated with exactly the same operations.

After the query hypervector is generated, the distance between the query vector and the class vectors must be determined. As shown in Fig.~\ref{fig:overview_hdc}-IV and mentioned in Section~\ref{sec:intro}, this is done by calculating the Hamming distance between the input vector and each of the 22 class vectors $d_h(\vec{\mathcal{T}}, \vec{\mathcal{L}}) = cnt_p(\vec{\mathcal{T}} \oplus \vec{\mathcal{L}})$. 
The Hamming distance is computed by performing an element-wise \texttt{XOR} operation followed by a population count on the resultant vector. As a final step, $\vec{\mathcal{T}}$ is classified into $\vec{\mathcal{L}_{\xi}}$ where $\xi = \text{argmin}_{i \in \{1,\dots,22\} }\left(d_H\left(\vec{\mathcal{T}},\vec{\mathcal{L}}_i\right)\right)$.

This method is based on the fact that the language vectors lie in a linear space that is spanned by a unique N-gram distribution of the associated language. 
The class vector with the closest N-gram distribution has the smallest distance to the input vector and represents the resulting language.
\begin{figure}[bp]
\centering
\includegraphics[scale=0.30]{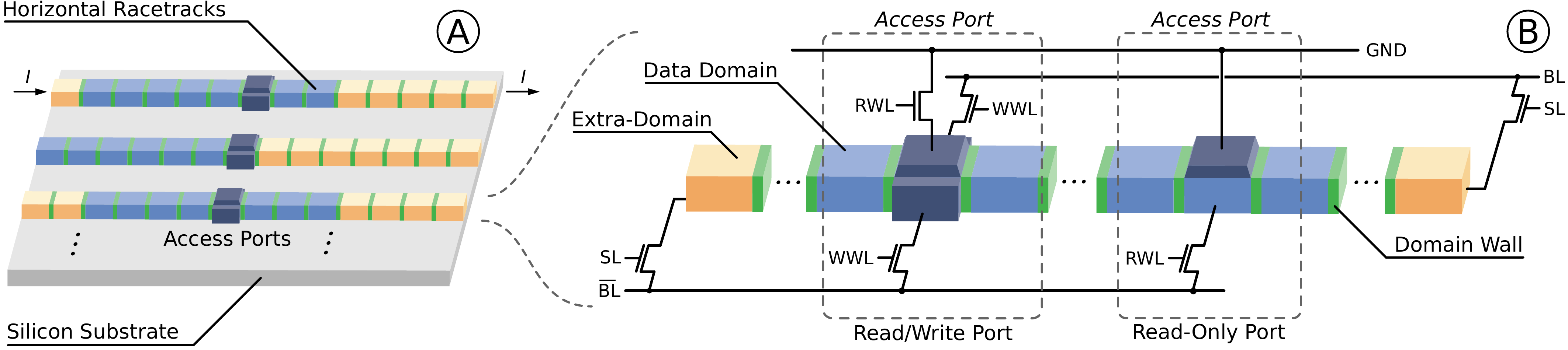}
\caption{\ac{RTM} nanowire structure (A) and anatomy(B).}
\label{fig:RTM}
\end{figure}

\subsection{Racetrack Memory}
\label{subsec:rtm}
The basic storage unit in racetrack memory is a 
magnetic nanowire that can be grown vertically or horizontally on a silicon wafer, as shown in Fig.~\ref{fig:RTM}. The nanoscale magnetic wires, also referred to as tracks or racetracks, can be physically partitioned into tiny magnetic regions called \emph{domains} that are delineated by \acp{DW} wherever the magnetization changes. This magnetization direction can be based on either in-plane ($\pm$ X) or perpendicular ($\pm$ Z) magnetic anisotropy. The state of any given domain exhibits a different resistance when it is parallel/antiparallel to a fixed reference domain, which can be interpreted as bits representing 1s and 0s~\cite{blaesing_2020}. Generally, each track in \ac{RTM} has its associated \acp{AP} and can store $K$ bits delineated by $K-1$ physical notches along the nanowire, where $K$ can be up to 128. The number of \acp{AP} per nanowire is usually less than the number of domains due to the larger footprint of the \acp{AP}~\cite{zhang2012perpendicular}. This mismatch in the number of domains and \acp{AP} leads to compulsory \emph{shifts}, \textit{i.e.}, each random access requires two steps to complete: \raisebox{.5pt}{\textcircled{\raisebox{-.7pt} {1}}} {\em shift} the target domain and {\em align} it to an \ac{AP} and \raisebox{.5pt}{\textcircled{\raisebox{-.7pt} {2}}} apply an appropriate voltage/current 
to {\em read} or {\em write} the target bit.

Shifting is conducted by passing spin-polarized current along the nanowire from either an access point or an endpoint to another access or endpoint; sufficient densities of spin-polarized current can overcome a potential well (``pinning'') created at notches and in turn advance all the domain walls toward the next notch position. This inherent behavior of \ac{RTM} can be imprecise, generating what is known as a ``shifting fault'' in the literature. Several solutions have been proposed to mitigate this fault mode~\cite{ollivier2019dsn, hifi, archer2020foosball}. 
Due to shifting, the access latency of \ac{RTM} is limited by the velocity with which domains move within the nanowire as well as the amount of shifts. The maximum number of domains per track depends on device parameters, but considering the user/application requirements and the number of \acp{AP}, the number of \emph{addressable} domains per track varies to accommodate shifting each addressable domain to align with any port. 

Fig.~\ref{fig:RTM} depicts the major components of an \ac{RTM} nanowire and its access circuitry. The blue domains represent the actual data stored in memory. The yellow domains are extra domains used to prevent data loss while shifting domain walls (and the data between them) along the nanowire. The dark blue elements and the connected access transistors form read-only or read-write ports.  A read-only port has a fixed magnetic layer, indicated in dark blue, which can be read using \texttt{RWL}.  The read-write port is shown using shift-based writing~\cite{DWM_Tapestri} where \texttt{WWL} is opened and the direction of current flows between \texttt{BL} and \textoverline{\texttt{BL}}.  Reading is conducted from \textoverline{\texttt{BL}} through the domain and \texttt{RWL} to \texttt{GND}. 

\begin{figure}[bp]
\centering
\includegraphics[scale=0.9]{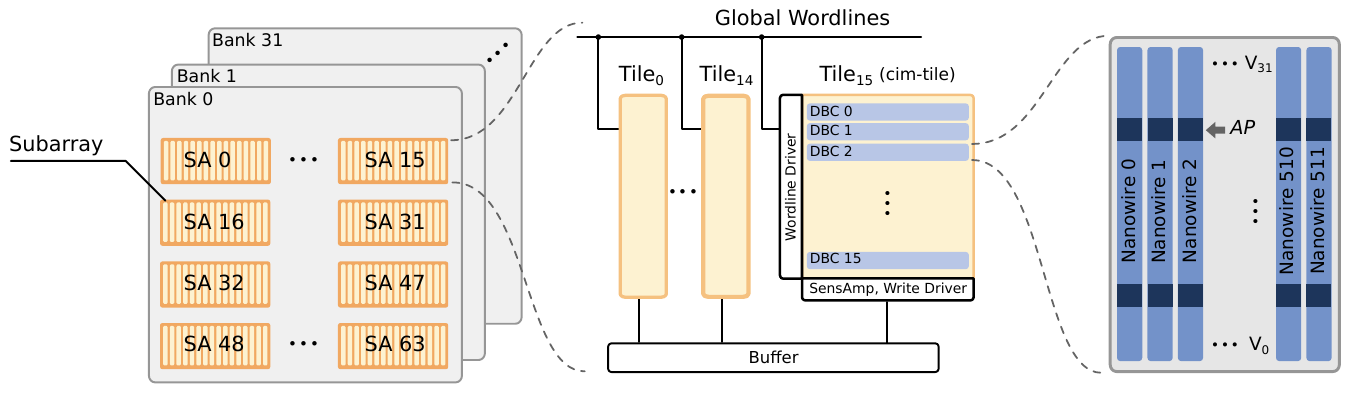}
\caption{\ac{RTM} organization. SA stands for subarray, \acp{DBC} for domain wall block cluster, \ac{AP} for access port, and SensAmp for sense amplifier.}
\label{fig:RTM_array}
\end{figure}

Similar to contemporary memory technologies, \ac{RTM} cells are grouped together to form a 2D memory array. 
To minimize the integration complexity, we deliberately conserve a DRAM hierarchical organization consisting of banks, subarrays, and tiles, as shown in Fig.~\ref{fig:RTM_array}.
As illustrated, the basic building block of the \ac{RTM} array is a group of $T$ nanowires and is referred to as a \acf{DBC}~\cite{tapecache, shiftsreduce}. A \ac{DBC} therefore can accommodate $K$ $T$-bit memory objects. Data in a \ac{DBC} is distributed across nanowires, which facilitates parallel access of all bits belonging to the same data word. Access ports of all $T$ tracks of a \ac{DBC} point to the same location and domains can be moved together in a lock-step fashion. For our proposed system, we use $K=32$ and $T=512$, the standard cache line size, as shown in Fig.~\ref{fig:RTM_array}. 
Note that for simplicity, we do not show the overhead domains in Fig.~\ref{fig:RTM_array} and $K$ refers to only addressable domains in the nanowires. We assume 16 \acp{DBC} per tile, 16 tiles per subarray. Furthermore, we assume a single \ac{CIM} tile or \emph{cim-tile} per subarray, capable of performing in-\ac{RTM} computations (see Section~\ref{sec:cim}).

~\\
\textbf{\ac{RTM} strengths, challenges and developments: }
Table~\ref{tab:memcomp} provides a direct comparison of \acp{RTM} to other memory technologies. \ac{RTM} offers high-performance SRAM comparable latency with extremely low leakage power and higher write endurance compared to other non-volatile memory technologies. However, due to the device's sequentiality, \ac{RTM} access latency and energy consumption depend on the number of required shift operations. In the worst case, the \ac{RTM} access latency can be 25.6$\times$ higher compared to an iso-capacity SRAM~\cite{stag}. In addition, shifts can also incur position and alignment faults. A number of solutions have been proposed to optimize \ac{RTM} performance through shift minimization~\cite{blaesing_2020}.  Additionally, solutions have been proposed to detect and correct \ac{RTM} misalignments~\cite{ollivier2019dsn,hifi}.

In recent years, \acp{RTM} have seen fundamental breakthroughs in device physics. In the earliest version of \ac{RTM}~\cite{stuart1.0}, controlled movement of domain walls in the nanowires was not only challenging but also extremely slow. Later, the same authors demonstrated accurate movement of domain walls with up to 10$\times$ higher velocities~\cite{stuart4.0}. More recently, the field-driving magnetic domain wall mobility has remarkably enhanced to 20 km/sT~\cite{dwspeed}, more than $20\times$ faster compared to the previous version or a two-order of magnitude improvement over the original prototypes. Similarly, moving domain walls in ferromagnetic materials with an exchange coupling torque~\cite{blasing2018exchange} has shown promise to reduce the critical current density to reduce shift energy. The data access devices, \acp{MTJ}, have also attracted significant interest and have observed considerable improvements in performance and thermal stability by employing different materials (\textit{e.g.,} MgO as a tunneling barrier) and adopting different switching mechanisms (such as spin-orbit torque instead of spin-transfer torque). These newer \acp{MTJ} allow for ultrafast magnetization switching, in sub-ns, with extremely low incident power~\cite{spindevices}. 

\begin{table}[tb]
\centering
\caption{Comparison of \acp{RTM} with other memory technologies~\cite{blaesing_2020}}
\centering
\label{tab:memcomp}
\scalebox{0.81}
{%
\begin{threeparttable}
\begin{tabular}{|c|c|c|c|c|c|c|}
\hline & SRAM 	& DRAM	& STT-MRAM	& ReRAM	& PCM	& RaceTrack 4.0\\  \hline
Cell Size ($F^2$) 	& 120-200 		& 4-8	& 6-50	& 4-10		& 4-12		& $\leq $ 2\\  \hline
Write Endurance 	& $\geq$ $10^{16}$ 	& $\geq$ $10^{16}$	& 4 X $10^{12}$		& $10^{11}$	& $10^9$ & $\geq 10^{16}$	\\  \hline
Read Time (ns) 		& 1--100 		& 30		& 3--15		& 10--20	& 5--20	& 3--250\tnote{\textdagger}   \\  \hline
Write Time (ns) 		& 1--100 		& 30		& 3--15     & 20		& $>$30	& 3--250\tnote{\textdagger} \\  \hline
Write Energy 	                    & Low 		    & Medium	& High	 & High		& High		 & Low\\  \hline
Read Energy 	                    & Low 			& Medium	& Medium	& Medium		& Medium		& Low \\  \hline
Leakage Power 	& High 	& Medium	& Low		& Low & Low	& Low \\\hline
Retention Period&Voltage-dependent&64--512ms&Variable& Years	& Years	& Years \\  \hline
\end{tabular}
\begin{tablenotes}
\item [\textdagger]including shift latency
\end{tablenotes}
\end{threeparttable}
}
\end{table}

~\\
\textbf{Transverse Read Operation in \ac{RTM}: } 
The \acf{TR} operation is an alternate access mode which conducts reads \emph{along} the nanowire rather than across it~\cite{ollivier2019dsn}.  By  applying a sub-shift-threshold current 
at an \ac{AP}, and performing a normal read at the next nearest \ac{AP} (for example, between the two access ports in Fig.~\ref{fig:RTM}), it is possible to detect how many of the domains between the ports are in a particular magnetic orientation.  The resultant magnitudes of the difference of resistances are small compared to the normal access mode, which limits how many domains can be accurately read in this manner without inadvertently shifting the domain walls.  However, using a \ac{TRD} of five domains can reliably produce a count of domains which are in either magnetic orientation~\cite{roxy2020novel}.


Prior work used this count to detect misalignment when shifting nanowires~\cite{ollivier2019dsn}, but this count can also be used to conduct bitwise logical operations on the data within the \ac{TRD}~\cite{CORUSCANT}.  Using a level-detecting sense amplifier, we can detect different voltage thresholds when $0,1,...,n$ bits are set, where exceeding any given threshold implies that all lower thresholds are also exceeded.  For example, if a TR is conducted across four words at a specific bit position in a nanowire, we derive logical \texttt{OR} if any of the thresholds are exceeded, logical \texttt{AND} if the threshold for four bits is exceeded, and \texttt{XOR} $\iff$ the threshold exceeded $\in \{1,3,5\}$.  For a fixed TR distance, these levels can be be used to realize carry-sum operations which can be composed to realize addition and multiplication~\cite{CORUSCANT}.  In the next section we show how a modified version of these level operations combined with handful of additional CMOS logic gates can be used to implement the fundamental HDC operations.

\section{Enabling Computation in Racetrack Memory}
\label{sec:cim}
This section presents the extensions to the cim-tile circuitry that enable in-place logical operations and counting in RTMs. 

\subsection{Logical Operations in RTM}
\label{subsec:RTMLogic}
Similar to~\cite{rahimi_ISLPED_16, rahimi_TCS_17}, we use the binary spatter-coding~\cite{BSC} framework that has four primary operations, \textit{i.e.}, \texttt{XOR} and circular shift operations for binding, the majority for bundling, and \texttt{XOR} for the similarity check as described in Section~\ref{subsec:hdc_bg}. 

\begin{figure}[bp]
\subfloat[Cim-tile architecture for in-place permutation and logical operations.  Additional logic relative to non-cim tiles is shown in red. `S' is sense amplifier and `SR' represents shift right.\label{fig:CIM-Architecture}]{%
  \includegraphics[width=0.45\textwidth]{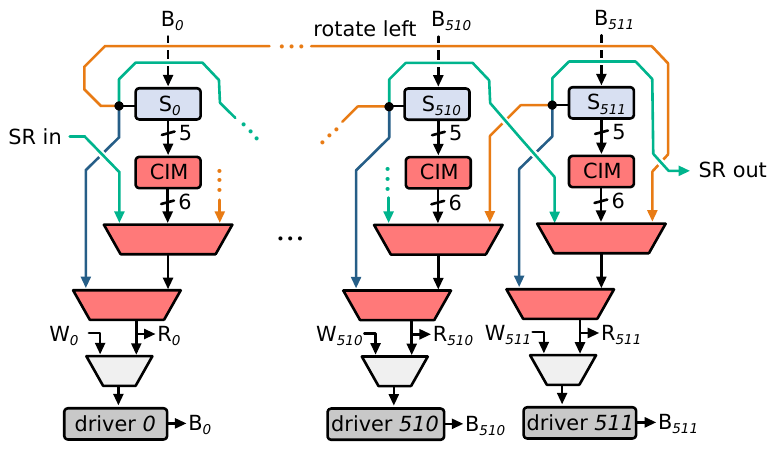}
}
\hfill
\subfloat[CIM block gates for logical operations.  Inputs i:j indicate that the reference voltages used to distinguish between i and j ones was exceeded.\label{fig:CIM-logic}]{%
  \includegraphics[width=0.45\textwidth]{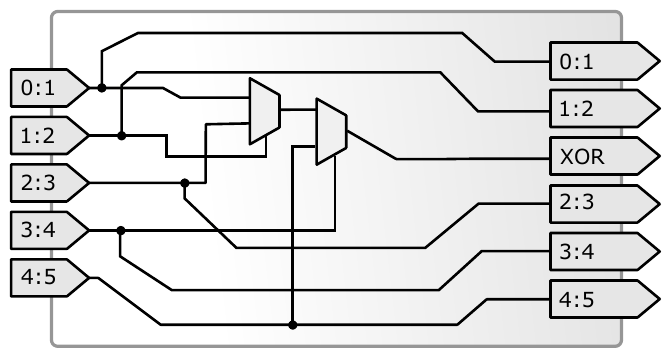}
}
\caption{Cim-tile architecture.} 
\label{fig:CIM-tile}
\end{figure}

To implement these operations in RTM, HDCR exploits the nanowires' properties and modifies the peripheral circuitry in selected RTM tiles (see Fig.~\ref{fig:RTM_array}), referred to as compute-in-memory tiles
. Concretely, one tile ($T_{15}$) in each subarray is designated as a cim-tile. Fig.~\ref{fig:CIM-Architecture} shows the necessary support circuitry similar to~\cite{CORUSCANT}, with the logic required for compute-in-memory operation outlined in red.  Sense amplifiers ($S_i$) shown in blue are aligned with access points at bitline $B_i$ to conduct either a normal read at that bit position, or a TR as described in Section~\ref{sec:background}.  During a TR operation, the sense amplifier outputs five bits indicating the five possible reference thresholds corresponding to a particular count of 1s between the access port at $B_i$ and another access port at a $TRD=5$ distance in the same nanowire. 
For example, 2:3 indicates that the voltage threshold between 2 and 3 ones was exceeded, indicating that at least 3 ones exist in the TR. 
To realize TR-based computations, we introduce the CIM block as shown in Fig.~\ref{fig:CIM-logic}.  Based on the thresholds representing the count of ones in the TR, and \texttt{XOR} is high when only the threshold for 0:1 is high or 2:3 is high with 3:4 being low, or when 4:5 is high.  The results of all operations are output simultaneously, to be selected using the multiplexer immediately below the CIM blocks.

\begin{figure}[!bp]
\subfloat[RTM counter overview.\label{fig:counter-overview}]{%
  \includegraphics[width=0.95\textwidth]{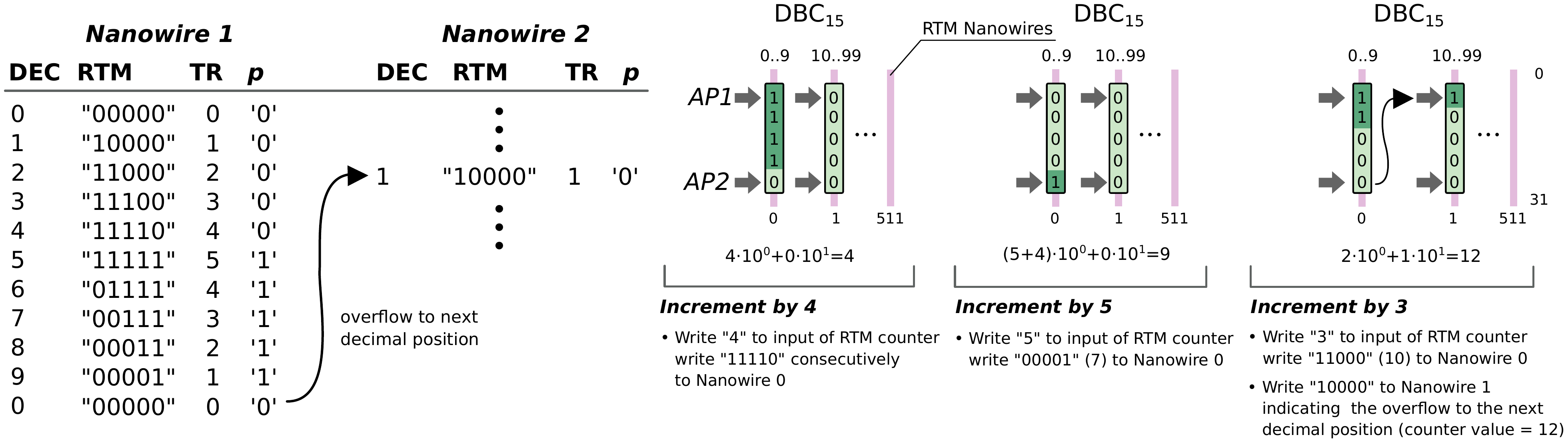}
}
\vfill
\subfloat[RTM counter implementation.\label{fig:counter-imp}]{%
  \includegraphics[width=0.9\textwidth]{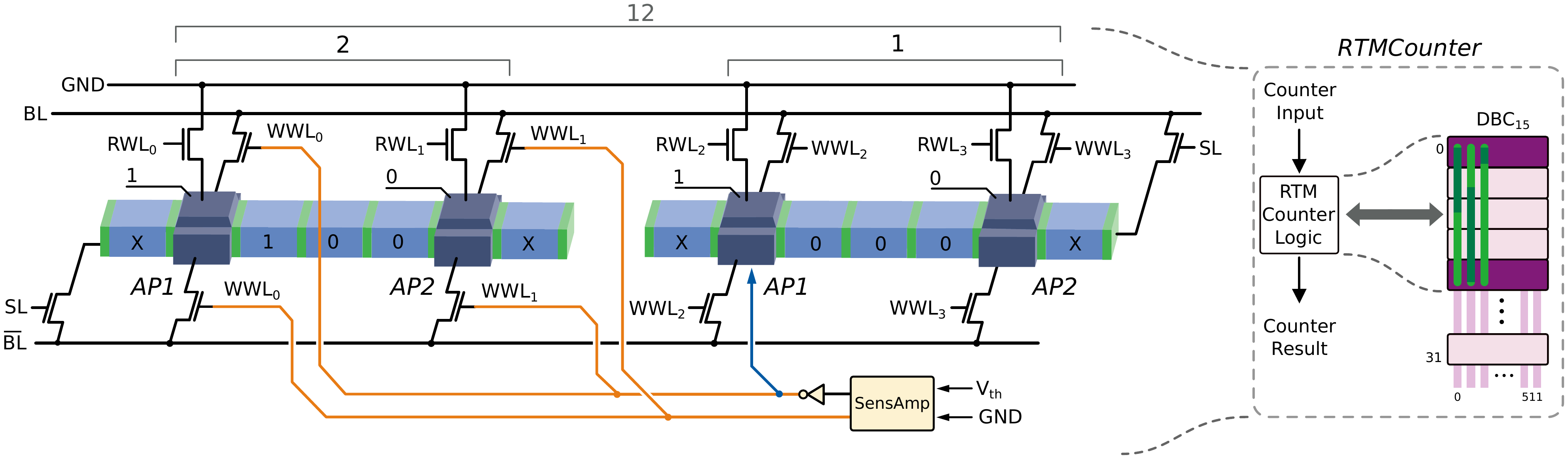}
}
\caption{RTM counter: overview and details.}
\label{fig:RTMCounter}
\end{figure}

During a normal read operation, each sense amplifier outputs the value of the single bit position directly beneath the access port.  This output bypasses the CIM tile and feeds directly to the first row of multiplexers to enable a fast read path. This same read path for bit line $B_i$ is routed to the multiplexer for the prior bit line $B_{i-1}$ and the subsequent bit line $B_{i+1}$, shown with orange and turquoise arrows, respectively. These paths enable circular shifting (permutation) of words by one bit position at a time. Together with the six outputs of the CIM block, the topmost row of multiplexers selects from eight operations on the input data.  The second row of multiplexers from the top is added to select from the CIM/shifting data path or the direct read path.  The final row of multiplexers and the writeback drivers are identical to the architecture of the non-cim tiles; data for writeback can be fed in from local row buffers $W_i$, or read from the current tile to move data or write back the result of a cim operation.


Operating this circuitry requires a new pseudo-instruction in the ISA called \emph{cimop}.  Each cimop instruction consists of a source address (\emph{src}), indicating which data to align to the access ports, a \emph{size}, indicating the number of nanowires to be included in the TR operation, and \emph{op}, which selects the cim operation from the topmost row of multiplexers.  Note that this psuedo-instruction entails some primitive operations to conduct the alignment and pad operands for sizes less than the TRD.  We assume that these primitive operations are scheduled by the compiler and conducted by the memory controller.

\subsection{Counting in RTM}
\label{subsec:counter}
Fig.~\ref{fig:counter-overview} presents an overview of the proposed in-RTM counter. It combines the TR operation in the RTM nanowire with the basic read/write operations to realize counters. The RTM nanowires used for counters must be equipped with two read-write APs, necessary for the TR operation. For a base$_{2\cdot X}$ counter, the two access ports in the nanowires must be $X-2$ domains apart, \textit{i.e.}, the TRD in the nanowire must be $X$.

In HDCR, we prefer decimal counters for the majority operation and the population count. As such, we use $X=5$, delimited by APs in dark blue in Fig.~\ref{fig:counter-imp} and with arrows in Fig.~\ref{fig:counter-overview}. Note that each nanowire in the RTM counter only uses the domain between the access ports and the number of nanowires in the counter are defined by the counter size. For instance, in a decimal counter, \textit{i.e.}, $X=5$, a single nanowire can only count between 0 and 9 (see Fig.~\ref{fig:counter-overview}). If we want to count from 0 and 99, the RTM counter requires at least two nanowires. In general, for a decimal counter having size $C$, an RTM counter requires at least $\lfloor \log_{10}(C) \rfloor+1$ nanowires.   

The RTM counter operates using the same principle as a Johnson counter.  Let us assume a two-nanowire decimal counter that can count up to 99 and is initially set to 0 (see Fig.~\ref{fig:counter-overview}).
The counter value at any instant in time is determined by the number of 1s between the APs and the state of bit $P$, the bit under AP2, \textit{i.e.}, the right AP in Fig.~\ref{fig:counter-imp}. The bit $P$ determines if the counter is in the first or second half of counting, in this case between 0-4 or between 5-9. For the decimal value 0, the $X$ bits are all filled with 0s and hence the bit $P$ is zero. If we want to increment the counter by four, for instance, four 1s need to be shifted under AP1, as shown in Fig.~\ref{fig:counter-overview}. To count beyond 5, \textit{i.e.}, when all bits between APs including the $P$ bit are 1, 0s are shifted under AP1. The decision to shift a 1 or a 0 under AP1 is controlled by the $P$ bit position: when $P=0$, we interpret the counter value as the count of ones between access points, and when $P=1$, we interpret the counter value as ten minus the count of ones  (or five plus the count of zeros) between access points. 
To realize this behavior, toggling the value of $P$ also toggles the value pushed into the nanowire when the counter is incremented, 
as shown for the decimal value 12 in Fig.~\ref{fig:counter-overview}. The table of Fig.~\ref{fig:counter-overview} represents all TR and $P$ combinations and their associated values. 


The RTM counter requires nanowires in DBCs to be shifted independently. This drastically increases the shift controller complexity since each nanowire AP position needs to be stored and controlled independently instead of a single position per DBC (512 nanowires).
In order to reduce this impact on the nanowire shifting logic, we also used the notion of \ac{TW}~\cite{CORUSCANT}.
Traditionally, to perform a shift based write under the left AP on Fig.~\ref{fig:counter-imp}, \texttt{RWL}$_0$ and one \texttt{WWL}$_0$ would be closed, the current flows through the fixed layer, one domain and then go to the ground, writing a new value and erasing the previous value under the left AP. However, by closing one \texttt{WWL}$_0$ and \texttt{RWL}$_1$, while sending a higher current density, our design 
can perform a write operation and perform a partial shift along the nanowire rather than between the fixed layer and ground. We called it partial (\textit{i.e.,} segmented) shift since only the bits between the heads are shifted. Thus, a TW from the leftmost AP writes a value under that AP, and shifts the remaining bits between the APs to the right, erasing the bit that was under the right AP. 


In the next section, we use these in-RTM compute-in-memory concepts and present our proposed architecture for HDCR. Further, we explain how the cim-tile operations implement each of the fundamental HDC operations.



\section{HyperDimensional Computing in Racetrack Memory}
\label{sec:HDCInRacetrack}

This section presents the implementation details of the proposed HDCR. 
It provides an overview of the overall system and explains the individual modules and their system integration.

\begin{figure}[tbp]
\subfloat[HDCR overview\label{fig:arch-overview}]{%
      \includegraphics[scale=0.98]{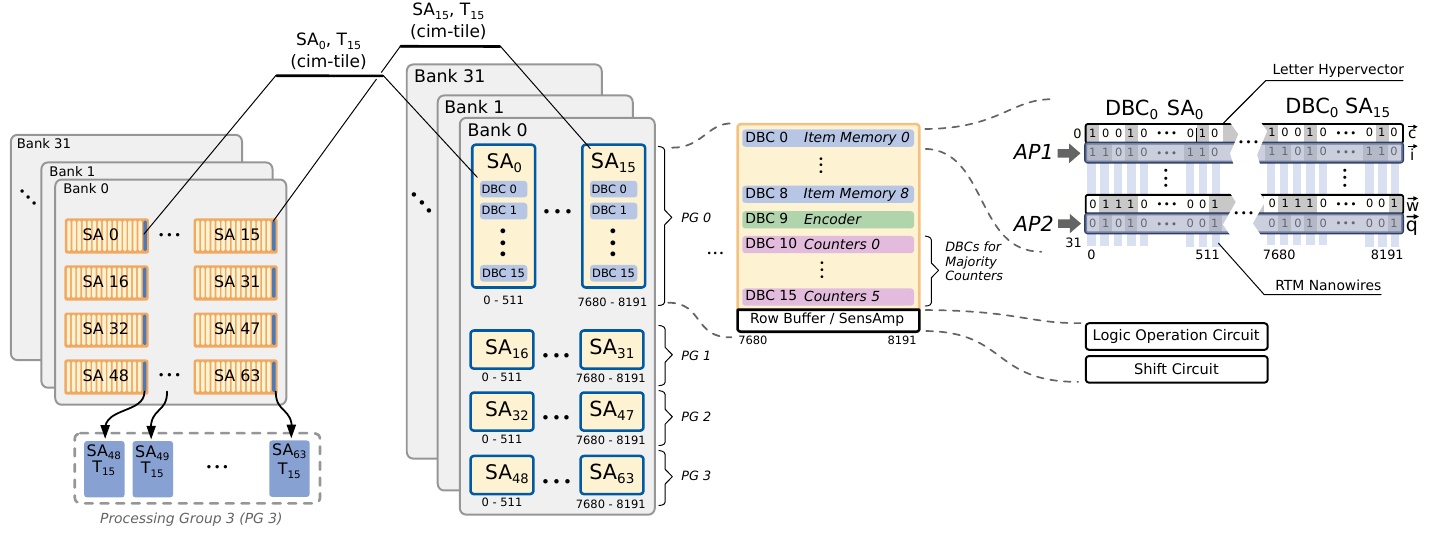}
     }
    \vfill
     \subfloat[HDCR workflow\label{fig:overview}]{%
      \includegraphics[scale=.78]{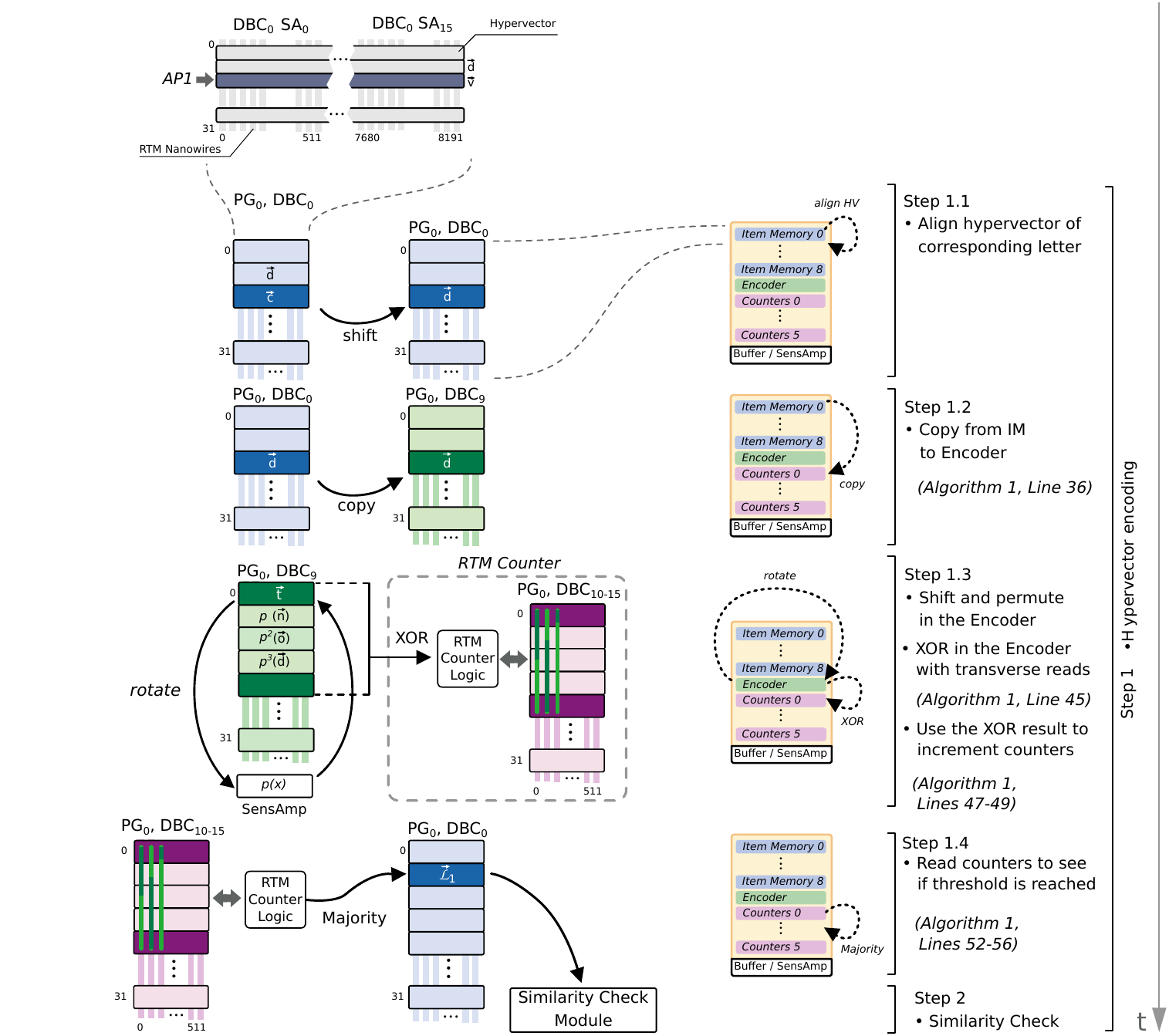}
     }
     \caption{An overview of the HDCR. The figure shows hypervectors' mapping to cim-tiles and provides detail of the individual operations in HDCR. Note that all tiles shown in the figure are cim-tiles.}
     \label{fig:overall}
\end{figure}
\begin{algorithm}[bth]
\caption{HDC Procedures}
\begin{algorithmic}[1]

\State \algorithmicvariables: $V_z \gets \emptyset, \theta, AM, THR$
\State  \Comment {$\theta$ = item memory, $AM$ = Associative memory, (cf. Section~\ref{subsec:use_case})}

\begin{multicols}{2}
{

\Function{HDC\_Train}{$LS, \theta$}
\State \Comment{LS: List of Lang strings for training}
    \ForAll{$L_i \in LS$}
        \State $\vec{\mathcal{L}_i} \gets \textsc{Encode(}L_i\text{)}$
        \State Store $\vec{\mathcal{L}_i}$ in AM
    \EndFor
\State
	\Return $AM$
\EndFunction

\vspace{0.6em}
\Function{HDC\_Classify}{$L, \theta, AM$}
\State \Comment{$L$: Text string to be classified}
   \State $\vec{\mathcal{T}} \gets \textsc{Encode(}L\text{)}$
    \State $\text{LangLabel} \gets \textsc{Sim\_Check(}\vec{\mathcal{T}}\text{)}$
    \State \underline{Display:} $L$ is \emph{LangLabel} language.
\EndFunction

    


\vspace{0.6em}
\Function{$\rho$}{$\vec{e}$}
    \State $\vec{\eta} \gets [], \vec{\psi} \gets []$
    
    \State PG\_size $\gets \frac{dim(\vec{e})}{T}$ \hspace{1em}\Comment{$T = 512$}
    \For{$Itr \gets 0$ to PG\_size}
        \State \Comment{Rotate left within each SA}
        \State $\vec{\eta} \gets$ \emph{rol }( $\vec{e}_{[512\cdot Itr] : [512 \cdot (Itr+1)] -1}$)\label{line:concatstart}
        \State \Comment{Concatenate rotated chunks}
        \State $\vec{\psi}_{[512\cdot Itr] : [512 \cdot (Itr+1)] -1} \gets \vec{\eta}\label{line:concatend} $
    \EndFor
\State
	\Return $\vec{\psi}$
\EndFunction

\vspace{0.5em}
\Function{Sim\_Check}{$\vec{\mathcal{T}}$}
	\ForAll{ $\vec{\mathcal{L}_i} \in AM$} 
	    \State \Comment{Implemented with TRs (cf. Sec~\ref{subsec:inference})}
        \State $d_H(\vec{\mathcal{T}}, \vec{\mathcal{L}_i}) \gets$ Hamdist$(\vec{\mathcal{T}}$, $\vec{\mathcal{L}_i})$
    \EndFor\label{alg:LoopEnd1} 
    \State \Comment{Implemented at the MemControl level}
    \State $\xi = \text{argmin}_{i \in \{1,\dots,22\} }(d_H(\vec{\mathcal{T}},\vec{\mathcal{L}}_i))$
\State
	\Return $\text{Label of language class }\vec{\mathcal{L}_\xi}$
\EndFunction

\Function{Encode}{String L}
\State $\vec{v_0} = \vec{v_1} = \vec{v_2} = \vec{v_3} \gets 0$
\State $N \gets 4, D \gets 8192$
\State $charCount \gets 0$
\State $counters \gets 0$ \Comment{$D$ counters in total}

	\ForAll{ $c_i \in L$} 
        \State $\vec{c_i} \gets \theta(c_i)$   \Comment{Read HV from IM} \label{line:copy} 
        \State\Comment{Rotate HVs in the N-gram}
        \State $\vec{v_3} \gets \rho(\vec{v_2})$\label{line:permutestart}
        \State $\vec{v_2} \gets \rho(\vec{v_1})$
        \State $\vec{v_1} \gets \rho(\vec{v_0})$\label{line:permuteend}
        \State $\vec{v_0} \gets \vec{c_i}$
        
        \State $charCount \gets charCount + 1 $
        \If{$charCount \ge N$}
            \State \Comment{XOR with a TR operation}
            \State $\vec{\phi} = \vec{v_0} \oplus \vec{v_1} \oplus \vec{v_2} \oplus \vec{v_3}$ \label{line:xor}
            \State \Comment{Push counters at all bit positions}
            
            \For{$Itr \gets 0 \text{ to } D$} \label{line:incstart}
                \If{$\vec{\phi}_{Itr} == 1$}
                     \State counters$_{Itr}++$
                \EndIf
            \EndFor \label{line:incend}
             
        \EndIf
    \EndFor\label{alg:LoopEnd} 
    
    \State 
    \State \Comment{Check all counters' state against THR}
    \For{$Itr \gets 0 \text{ to } D$}\label{line:thrstart}
        \If{counters$_{Itr} > THR$}
             \State $\vec{\mathcal{T}_{Itr}} \gets 1$
             \Else
             \State $\vec{\mathcal{T}_{Itr}} \gets 0$
        \EndIf
    \EndFor \label{line:thrend}
\State
    \Return $\vec{\mathcal{T}}$
    \item[]
\EndFunction

}\end{multicols}
\end{algorithmic}

\label{alg:hdc}
\end{algorithm}

\subsection{Overview}
\label{sec:overview}
Fig.~\ref{fig:overall} presents an overview of the proposed in-RTM HDC system. As explained in Section~\ref{subsec:hdc_bg}, the 27 hypervectors of the input letters are initially mapped to the item memory, 9 DBCs in each subarray as shown in Fig.~\ref{fig:arch-overview}. Note that for simplicity, we only show the cim-tiles in the subarrays. For the encoding operation, the hypervectors in the item memory are loaded into the encoder module. This requires the hypervectors in the item memory to be shifted and aligned to the port positions in their respective DBCs (Step 1.1 in Fig.~\ref{fig:overview}). Subsequently, HDCR copies the hypervectors to the encoder module implemented in DBC$_{9}$ of the subarray (see Step 1.2 in Fig.~\ref{fig:overview} and Line~\ref{line:copy} in Algorithm~\ref{alg:hdc}). HDCR then permutes the hypervectors in the encoder module (see Lines~\ref{line:permutestart}-\ref{line:permuteend} in Algorithm~\ref{alg:hdc})  and performs the XOR operation to generate their N-gram hypervector (see Step 1.3 in Fig.~\ref{fig:overview} and Line~\ref{line:xor} in Algorithm~\ref{alg:hdc}). Since the N-grams represent $N$ contiguous characters in the input text, the encoder module produces a new N-gram hypervector for each new character in the text. Thus for an input text of $S$ characters, the encoder module generates $S-N+1$ hypervectors in total. 

For each new N-gram hypervector, the counters for each bit position implemented in DBCs$_{10-15}$ are incremented based on the XOR result (see Step 1.4, Lines~\ref{line:incstart}-\ref{line:incend} in Algorithm~\ref{alg:hdc}). The counting module performs the majority operation on all N-gram hypervectors and generates a single hypervector based on the final counters' state (Step 1.4). In the training phase of the HDC this generated hypervector represents a language class hypervector ($\vec{\mathcal{L}_i}$).  This is stored in the \ac{AM}, and the process is repeated for all remaining languages. In contrast, during the inference phase, the resultant hypervector ($\vec{\mathcal{T}_i}$) represents the input text.  After generating this hypervector, it is passed on to the similarity search module in Step 2 to classify it into one of the language classes, as shown in Fig.~\ref{fig:overview}. In the following sections, we provide the implementation details of the individual modules.


\subsection{Item Memory}
\label{sec:IM}
The HDC framework operates on $D=8192$ bit wide binary vectors. Since our DBCs are only 512 bits wide, this requires dividing the hypervectors into 16 chunks of 512 bits each to store the complete 8192-bit hypervector. These chunks can be stored in DBC(s) of the same subarray, as we are doing in Section~\ref{subsec:inference}, or in the same DBC (\textit{e.g.,} $DBC_i$) across 16 different subarrays. However, for the encoder module in HDCR, to enable performing the TR operation in parallel across all 8192 bit-positions, the HV chunks need to be distributed across different subarrays, as shown in Fig.~\ref{fig:arch-overview}. This group of 16 subarrays sharing and manipulating chunks of the same hypervectors is referred to as a \ac{PG}. A \ac{PG} generates the output of a CIM operation on TRD hypervectors in a single cycle.

For the LR application, the item memory (IM) is composed of $27$ hypervectors (HVs), one for each character of the Latin alphabet plus the space character $\tau$ (see Section~\ref{subsec:use_case}).  
Since a DBC in our proposed system has 32 domains per nanowire, the 27 HVs can be stored in a single DBC (\textit{e.g.}, $DBC_0$) across all subarrays in a \ac{PG}. However, since each new character consumed from the input text accesses the IM to retrieve its corresponding HV, this tight packing of HVs in a single DBC can lead to a significant number of shift operations in RTM. 
In the worst case, access to the IM can incur $27 - TRD = 23$ shifts, which stalls the other modules in HDCR and substantially increases the overall runtime. To overcome this, HDCR dedicates 9 DBCs (see Fig.~\ref{fig:arch-overview}) to the IM and distributes the HVs in the IM such that accessing an HV requires at most one RTM shift.  That is, by placing each character HV directly at or adjacent to one of the two access ports, we can access the 18 HVs beneath the access ports without shifting, and the remaining 9 HVs by shifting by one position.

To efficiently map the character HVs into the IM, we profiled each language to rank the frequency of each character in our corpus.  The most frequently occurring characters are then placed directly under the access ports, and the remaining characters are distributed among the bit positions adjacent to the access ports.


\subsection{Encoding}
\label{subsec:encoding}
The encoder module transforms the entire language into a representative vector (see Section~\ref{subsec:hdc_bg}).
From the implementation perspective, the encoder module performs three major operations, \textit{i.e.}, binding, permutation and bundling (see Fig.~\ref{fig:overview}). In the following sections, we explain how these operations are implemented. 


\subsubsection{Binding and Permutation in RTM}
\label{sss:binding}
As explained in Section~\ref{subsec:hdc_bg}, the binding operation in HDC generates a new hypervector by \texttt{XOR}ing the permuted versions of the $N$ character hypervectors which form each N-gram in the input text.

Initially, all hypervectors of the respective N-gram are iteratively loaded into the encoder module \textit{i.e.}, DBC$_{9}$ (see step 1.2 in Fig.~\ref{fig:overview}). Depending on the HVs position in the IM, this may require a shift operation in RTM, as demonstrated in Fig.~\ref{fig:overview} (step 1.1). 
In the next step, the hypervectors are rotated by $M$ times, where the value of $M$ for a particular hypervector depends upon its position in the N-gram. This rotation is functionally equivalent to a bitwise circular shift, where the $M$ most significant bits overwrite the $M$ least significant bits after shifting the remaining $512-M$ bits left by $M$ bit positions. Note that this shifting is different from the RTM nanowire shift operation. In this case, the HV bit positions along the nanowire do not change, rather the HV representing the character is shifted across all nanowires it spans, using the peripheral circuitry in Fig.~\ref{fig:CIM-Architecture}. 
For instance, for the first N-gram ``dont'' in the running example, the hypervector $\vec{d}$ of the first character `d' is rotated by 3, the hypervector $\vec{o}$ is rotated by 2, the hypervector $\vec{n}$ is rotated by 1, and the hypervector $\vec{t}$ is taken unchanged. This is important for differentiating this permutation of these four characters from any other permutation.

To efficiently rotate a hypervector, which spans many DBCs, the \emph{rotate} control signal is enabled and a read operation is performed on all subarrays in a \ac{PG}. The resultant hypervector in the row-buffer is the rotated-by-one version of the original hypervector. A subsequent write command is issued to the RTM controller to update the new value in RTM. To perform a rotation by three, our RTM architecture will perform three rotated-by-one operations sequentially.

Note that rotating an entire \SI{8192}{\bit} HV in RTM requires considerable modifications to the RTM row buffer. The customization in Fig.~\ref{fig:CIM-Architecture} only allows rotating a 512-bit chunk of the HV, \textit{i.e.}, rotation at the granularity of the subarray. HDCR performs chunk-wise permutation on all subarrays in a \ac{PG} and concatenates the permuted chunks to generate the permuted HV, as demonstrated in Fig.~\ref{fig:overview} (Step 1.3) and Algorithm~\ref{alg:hdc} (Lines~\ref{line:concatstart}-\ref{line:concatend}). 
This chunk-wise rotation operation is reversible and the generated hypervector was empirically verified to not adversely impact the accuracy of the HDC framework.




Once the required $N$ hypervectors for a particular N-gram are loaded and $N-1$ (all but last) hypervectors are permuted, they are \texttt{XOR}ed together to generate the resultant N-gram hypervector ($\vec{\phi_i}$). As described in Section~\ref{subsec:rtm}, a TR operation and sense amplifiers detect how many ones exist between the TR access ports.  When exactly one, three, or five 1s are detected, the logic in Fig.~\ref{fig:CIM-logic} asserts the \texttt{XOR} output, representing the \texttt{XOR} of all TRD operands. 

This binding operation is performed iteratively for all N-grams in the input text. As the input text is consumed, each character hypervector in each N-gram is used at least 
$N$ times in different permutations to generate $N$ N-gram vectors. For instance, the hypervector $\vec{t}$ is used as-is to generate the first N-gram vector in the running example. However, for the second N-gram (``ont$\tau$'') vector, $\vec{t}$ is rotated by 1. Similarly, for third and fourth N-gram vectors, $\vec{t}$ is rotated by 2 and 3, respectively. Since the sequence of operations is known, we can reuse each permutation result in the next iteration to save execution cycles. 

\begin{algorithm}[bth]
\caption{Memory operations required for computing an N-gram HV}
\begin{algorithmic}[1]

\State  \Comment {$\vec{v_i} \text{, } i \in \{0,1,2,3,4\}$ represents HV stored in DBC locations 0,1,2,3,4, i.e., all five locations between APs (see Step 1.3 in Fig.~\ref{fig:overview})}

\State \Comment {At any time \underline{Shift} (if necessary) to align $\vec{c_i}$ to AP in IM}
\item[]
\State Algorithm Step: $\vec{v_1} \gets \rho(\vec{v_0})$ (see Line 40 in Algorithm~\ref{alg:hdc})
\State Memory operations:
\Statex (i) \underline{Read} $\vec{v_0}$ (with rotate signal enabled)
\Statex (ii) \underline{Write} the row buffer contents to lower access point (old \texttt{V}$_0$, new \texttt{V}$_1$)

\State Algorithm Step: $\vec{v_2} \gets \rho(\vec{v_1})$ (see Line 39 in Algorithm~\ref{alg:hdc})
\State Memory operations:
\Statex (i) \underline{Shift} down one position to align $\vec{v_1}$ to lower AP
\Statex (ii) \underline{Read} $\vec{v_1}$ (with rotate signal enabled)
\Statex (iii) \underline{Write} the row buffer contents to lower access point (old \texttt{V}$_1$, new \texttt{V}$_2$)

\State Clear old $\vec{v_3}$:
\State Memory Operations
\Statex (i) \underline{Shift} up by three positions to align $\vec{v_3}$ to upper AP while resetting row buffer
\Statex (ii) \underline{Write} the row buffer contents to upper access point (old \texttt{V}$_3$, new \texttt{V}$_4$)

\State Algorithm Step: $\vec{v_3} \gets \rho(\vec{v_2})$ (see Line 38 in Algorithm~\ref{alg:hdc})
\State Memory operations:
\Statex (i) \underline{Shift} up by one position to align $\vec{v_2}$ to AP
\Statex (ii) \underline{Read} $\vec{v_2}$ (with rotate signal enabled)
\Statex (iii) \underline{Shift} to align DBC location three to AP
\Statex (iv) \underline{Write} the row buffer contents to the DBC upper access point (old \texttt{V}$_2$, new \texttt{V}$_3$)

\State Algorithm Step: $\vec{v_0} \gets \vec{c_i}$ (see Line 41 in Algorithm~\ref{alg:hdc})
\State Memory operations:
\Statex (i) \underline{Shift} down by one position to align DBC new \texttt{V}$_0$ to lower AP and \underline{Read} $\vec{c_i}$
\Statex (ii) \underline{Write} the row buffer contents to the DBC \texttt{V}$_0$


\end{algorithmic}

\label{alg:permute}
\end{algorithm}

To accomplish this we leverage both upper and lower access points to align, read/shift into the row buffer, and then write back the rotated into the access points while minimizing alignment operations.  The detailed approach is described in Algorithm~\ref{alg:permute} referencing DBC locations from Fig.~\ref{fig:RTM_array} in the encoder DBC$_9$ shown in Fig.~\ref{fig:overall}.  Using the example, we first read $\vec{v_0}$ and rotate and then write it back to complete $p^1 (\vec{n})$.  We then align $\vec{v_1}$ with the lower access point to complete $p^2 (\vec{o})$.  We then align the outgoing $\vec{v_3}$ with the upper access point to reset it to zero.  We then align $\vec{v_2}$ with the upper access point to complete $p^3 (\vec{d})$ and then align the lower access point to write $\vec{t}$ from the IM.

As a result of the binding and permute operation, a new N-gram vector is generated and is consumed by the bundling unit, as explained in the next section. For the entire input text, a whole set of N-gram vectors is generated where each vector corresponds to an N-gram in the text. 
Recall, $V_z$ represents all N-gram vectors of the input text (see Section~\ref{sec:background})\footnote{$V_z$ is distinct from \texttt{V}$_{0..31}$, which represents logical locations in the DBC (see Fig.~\ref{fig:RTM_array}).}.
The bundling operation combines all elements in $V_z$ by taking the bit-wise majority on each bit position, as explained in Section~\ref{sec:background}.  In the next section, we discuss the implementation of bundling in HDCR.

\subsubsection{Bundling Operation in RTM}
\label{subsec:bundling}
Bundling in the HDC framework is a conjunctive operation that forms a representative vector for the set of N-gram hypervectors $V_z$ (see Section~\ref{subsec:hdc_bg}). Concretely, it computes a new hypervector $\vec{\Gamma}$ by adding all hypervectors in $V_z$, \textit{i.e.}, $\vec{\Gamma} = \sum_{\vec{\Phi} \in V_z}\vec{\Phi}$. Each component in $\vec{\Gamma}$ is then compared to a fixed threshold to make it binary
, \textit{i.e.}, $\forall  i \in \{1,2,\dots 8192\}, \vec{\mathcal{T}_i} = \beta_i, \text{and}$
\[
    \beta_i= 
\begin{cases}
    1,  & \text{if } \vec{\Gamma_i} > \text{threshold}\\
    0,              & \text{otherwise}
\end{cases}
\] (see Algorithm~\ref{alg:hdc}, Lines~\ref{line:thrstart}-\ref{line:thrend}). 
The threshold value for binary hypervectors is typically the greatest integer less than 0.5 times the number of elements in $V_z$.
For instance, for $|V_z| = 55$, the threshold value will be $\lfloor{55\times0.5} \rfloor = 27$, which also means that the resultant hypervector $\vec{\mathcal{T}}$ is equivalent to the output of the \emph{majority} function, \textit{i.e.}, $\vec{\mathcal{T}} = \text{Majority}( \text{ }\vec{\Phi}, \text{ }\forall \vec{\Phi} \in V_z)$. 

HDCR uses RTM counters (see Section~\ref{subsec:counter}) for each bit position to implement the majority function for $|V_z|> \text{TRD}$. As shown in Fig.~\ref{fig:overview} (step 1.2-1.4), each subarray dedicates DBCs$_{10-15}$ for RTM counters. At each bit position in a \ac{PG}, the 6 nanowires in DBCs$_{10-15}$ are used to implement the counter for that particular position. With 6 nanowires, the RTM counters can count from $0$ to $10^6-1$, far more than what is required for the LR use case. For each new N-gram hypervector, HDCR updates all counters simultaneously based on the \texttt{XOR} result. Once a particular counter hits the threshold, it ignores subsequent incrementing. To simplify the thresholding, the memory controller can preset the state of the counter to $M-T$ where $M$ is the maximum value represented by the counter and $T$ is the desired threshold.  Thus, the thresholding does not require any additional logic and can be represented by the status of the $P$ bit of the most-significant digit of the counter.  

In our evaluated system, we have 128 \acp{PG} (see Section~\ref{subsec:setup}). To reduce the overall runtime, the input text is divided into 128 chunks, and each chunk is provided to a separate \ac{PG}. Once the computation in all \acp{PG} is finished, the majority output of all \acp{PG} is combined to make a single final vector. In the training phase of the HDC framework, this final computed hypervector represents the language (class) hypervector and is written to the \ac{AM} (same DBCs as for item memory, i.e., DBCs$_{0-9}$ but different positions). In inference, this hypervector is referred to as the query hypervector and is compared to all class hypervectors to infer the final result, as shown in Fig.~\ref{fig:overview} (step 2) and explained in the next section.

\subsection{Inference}
\label{subsec:inference}
The inference phase of the HDC framework uses the same encoding module to generate a query hypervector for the input text. Since the language class hypervectors are pre-generated in the training phase and are stored in the cim-tiles, classification is conducted by computing the Hamming distance of the query vector with all class vectors to find the closest match (see Section~\ref{subsec:hdc_bg}).

This similarity search is encapsulated in a module which performs three main operations. First, the query hypervector is \texttt{XOR}ed with all class hypervectors for bit-wise matching. Subsequently, the Hamming weight is computed by performing a population count of set bits within each of the computed hypervectors. Finally, the language with the minimum Hamming weight is inferred as the output. 

\begin{figure}[tbh]
\centering
\includegraphics[scale=0.9]{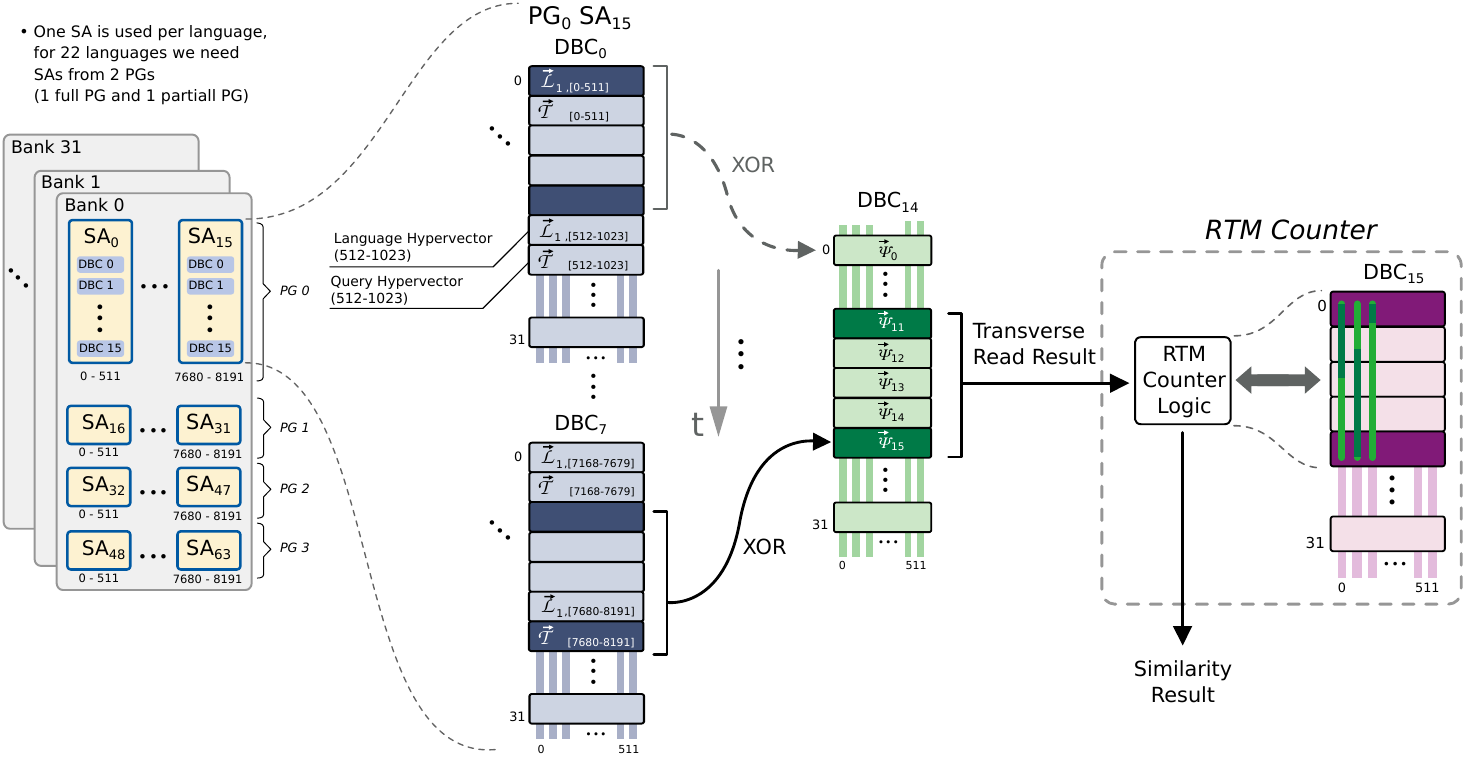}
\caption{Similarity search module}
\label{fig:search_module}
\end{figure}

From the implementation perspective, HDCR uses one subarray per language hypervector. For the 22 language hypervectors, HDCR uses 22 subarrays (2 \acp{PG}). 
As shown in Fig.~\ref{fig:search_module}, the language vectors in subarrays are stored across different DBCs of the same subarray, unlike the encoding module which stores hypervectors across different subarrays. The query vector is then written to all 22 subarrays to compute the Hamming weights independently.

The \texttt{XOR} operation generates 16$\times$512 bits for each language. In each subarray (for each language), the 16 chunks are processed sequentially, with each iteration producing one 512-bit chunk of the \texttt{XOR} operation in a single cycle, and then storing the results adjacent to one another in the same DBC (DBC$_{14}$ in Fig.~\ref{fig:search_module}) for the subsequent population count operation. 
For each of these 16 parallel \SI{512}{\bit} results, the TR operation sequentially performs the `1' counting in DBC$_{14}$. HDCR uses the TR result to shift bits in the RTM counter implemented in DBC$_{15}$, as shown in Fig.~\ref{fig:search_module}. Since the maximum count value in the similarity search module can be 8192, HDCR uses four nanowires for the RTM counter in this module. Note that, unlike the per-bit counting for the majority operation in the encoding module, the similarity search module uses a single RTM counter per DBC to find a single Hamming weight value per language. This necessitates the counters to be updated sequentially for all 512 TR outputs after each TR operation. 


Once the counting operations of the inference is done, the TR and $P$ values for all counters packed like in Fig.~\ref{fig:packing-results} and sent sequentially to the memory controller for final input language selection. 
\begin{figure}[h]
\smaller
\begin{tabular}{lccccccccc}
Index & 0..4 & 5 & 6 .. 10 & 11 & 12 .. 16 & 17 & 18 .. 22 & 23 & 24 .. 511\\
\hline
Value & \multicolumn{1}{|c}{$TR_{00}$ .. $TR_{04}$} & \multicolumn{1}{|c}{$P_0$} & \multicolumn{1}{|c}{$TR_{10}$ .. $TR_{14}$} & \multicolumn{1}{|c}{$P_1$} & \multicolumn{1}{|c}{$TR_{20}$ .. $TR_{24}$} & \multicolumn{1}{|c}{$P_2$} & \multicolumn{1}{|c}{$TR_{30}$ .. $TR_{34}$} & \multicolumn{1}{|c}{$P_3$} & \multicolumn{1}{|c|}{$\emptyset$ .. $\emptyset$} \\\hline
 
\end{tabular}
\caption{Example of packing TR and $P$ values from the counters into local subarray rowbuffer.}
\label{fig:packing-results}

\end{figure}
\section{Evaluation}
\label{sec:Evaluation}
This section explains our experimental setup, provides details on the dataset, and compares our proposed system to state-of-the-art solutions for performance and energy consumption. 
Concretely, we evaluate and compare the following designs.
\begin{itemize}
    \item \emph{HDCR:} Our proposed in-RTM HDC system.
    \item \emph{FPGA:} The FPGA based HDC system from ~\cite{rahimi_ISLPED_16}.
    \item \emph{PCM:} The in-PCM HDC implementation from ~\cite{inPCM_2020}.
    \item \emph{CPU:} For the sake of completeness, we also compare to a software/CPU control.
\end{itemize}

\subsection{Experimental Setup}
\label{subsec:setup}
As a target system, we consider an RTM-based 8GB main memory that consists of 32 banks, having 64 subarrays each. A subarray consists of 16 tiles composed of 16 DBCs, which are 512 bits wide and have 32 columns/data domains per racetrack. We assume two access ports per nanowire and an operating clock frequency of \SI{1000}{\mega\hertz}. The cim-tiles utilize a high throughput mode proposed in prior PIM work~\cite{deng2018dracc}.
The peripheral circuitry in cim-tiles does not affect the storage capability or otherwise prevent its use to store data beyond the marginal delay of a single multiplexer. The majority of the latency overhead results from the reducing the number of domains between the ports, from 16 to 5, which increases the average shift distance in the cim-tiles.
While the target technologies may be subject to different types of faults, the experiments here presume fault free operation to ensure a fair comparison, particularly with respect to PCM which has limited endurance.  However, HDCR is is compatible with previous reliability schemes proposed in the literature as DECC~\cite{ollivier2019dsn}, or Hi-Fi~\cite{hifi} and by employing these techniques the major fault mode of shift misalignment the intrinsic fault rate of circa $10^{-5}$ can be reduced to circa $10^{-20}$ with negligible performance penalty~\cite{hifi}. For the LR use case, the entire training and test data sets fit in RTM. However, since the proposed solution is generic and use case independent, the data sets can also be partially loaded into RTM as needed to accommodate larger inputs with the same size working set. 
The energy and latency numbers of the memory subsystem are estimated using the CIM architecture presented in~\cite{CORUSCANT}, the parameters from~\cite{DW-NN} and are shown in Table~\ref{tab:RTMValues}.

\begin{table}[tbh]
\centering
\caption{RTM latency and energy parameters}
\centering
\label{tab:RTMValues}
  \begin{tabular}{c|c}
\toprule
Domains per track & 32 \\
Tracks per DBC & 512 \\
Background power [\SI{}{\milli\watt}] & 212 \\
rRead energy [\SI{}{\pico\joule}]/bit & 0.5 \\
Shift energy [\SI{}{\pico\joule}]/bit & 0.3 \\
Shift latency [Cycle] & 1 \\
Read latency [Cycle] & 1 \\
Write latency [Cycle] & 1 \\
\bottomrule
\end{tabular}
\end{table}

~\\
\textbf{Baseline Systems:} For the FPGA design, we use the System Verilog implementation from~\cite{rahimi_ISLPED_16}. We synthesize the design on a Xilinx Virtex 7 FPGA (7vx1140tflg1930) using Vivado 19.2. The maximum clock frequency was \SI{80}{\mega\hertz} and the device utilization is 61\% and 23\%, for LUTs and flip flops, respectively. We get the throughput result from the post place \& route simulation, which was also used to record the switching characteristics of the design. The switching activity file is fed to the Vivado power estimator to get the overall energy consumption. 

For the CPU results, we use an Intel$^\text{\textregistered}$ Core(TM) i7-5650U CPU @ \SI{2.20}{\giga\hertz}, with \SI{8}{\giga\byte} RAM. We use the C libraries for the LR use case from~\cite{hdc_C}. 
For comparison with the PCM configuration, we used the numbers reported in~\cite{inPCM_2020}.

\subsection{Data Set}
The language training data is taken from a corpora~\cite{LR_training_corpus}, which contains sample texts in 22 languages. For inference, an independent data set from~\cite{LR_inference_corpus} is used, which comprises 1000 sentences per language. The training, respectively the derivation of the language hypervectors, was carried out with the entire training data set, which contains a text of 120000-240000 words per language. The classification and thus the evaluation of the accuracy is carried out on multiple instances of one sentence per language. Concretely, 1000 tests with one sentence per test are performed for each language. We implement both the training and the inference phases of the HDC framework and report the results in the following sections.

\subsection{Performance Comparison}

\begin{figure*}
\pgfplotsset{compat = newest}
\pgfplotsset{major grid style={dotted,aluminium2!50!black}}
\begin{tikzpicture}
\begin{axis}
[
    width=1\textwidth,
    ybar=1pt, 
    enlargelimits=0.035,
    enlarge y limits={upper, value=0.2},
    ylabel style={align=center},
    y label style={at={(-0.05,0.5)}},
    ylabel=Runtime (ns),
    legend style={draw=none, fill=none},
    bar width=3.5pt,
    legend columns=2,
    ymin=100000, 
    ymode = log,
    log basis y={10},
    height=0.27\textwidth,
    ymajorgrids=true,
    grid style=dashed,
    axis x line*=bottom,
    x tick label style={xshift=.4em,rotate=45,anchor=east},
    yminorticks=true,
    legend style={at={(1,1.2)},anchor=north east},
    xlabel={},
    every y tick scale label/.style={at={(yticklabel cs:1.2)}, anchor = north west, rotate = 0},
    symbolic x coords={afr, bul, ces, dan, deu, ell, eng, est, fin, fra, hun, ita, lav, lit, nld, pol, por, ron, slk, slv, spa, swe},
    xtick=data,
]


\addplot+ [blind_safe_one_scheme_three_colors] coordinates {
(afr,1120268  ) 
(bul,1268131 ) 
(ces,1144858  )
(dan,1485636 ) 
(deu,1357005 ) 
(ell,1520539 ) 
(eng,1214924 ) 
(est,1142261  ) 
(fin,1208138 ) 
(fra,1415764 ) 
(hun,1377901 ) 
(ita,1467752 ) 
(lav,1258952 ) 
(lit,1226659 ) 
(nld,1132938  ) 
(pol,1177833  ) 
(por,1484404 )
(ron,1436029 ) 
(slk,1232416 ) 
(slv,1344571 ) 
(spa,1581871 ) 
(swe,1111013  ) }; 

\addplot+ [blind_safe_two_scheme_three_colors] coordinates {
(afr,11390340  ) 
(bul,12896100 ) 
(ces,11637072  )
(dan,15109524 ) 
(deu,13797828 ) 
(ell,15463584 ) 
(eng,12352944 ) 
(est,11620452  ) 
(fin,12281712 ) 
(fra,14395968 ) 
(hun,14013288 ) 
(ita,14927676 ) 
(lav,12799968 ) 
(lit,12473820 ) 
(nld,11524572  ) 
(pol,11973948  ) 
(por,15092760 )
(ron,14598240 ) 
(slk,12532056 ) 
(slv,13671768 ) 
(spa,16084680 ) 
(swe,11296152  ) }; 


\legend{HDCR, FPGA}
\end{axis}
\end{tikzpicture}
\caption{Runtime of HDC training on different platforms.}
\label{fig:runtime_train}
\end{figure*}
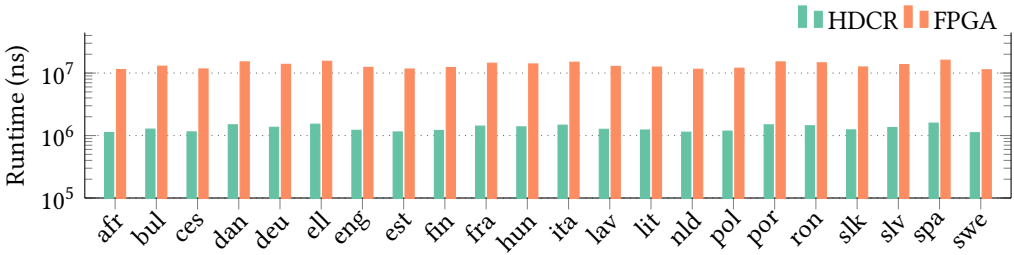
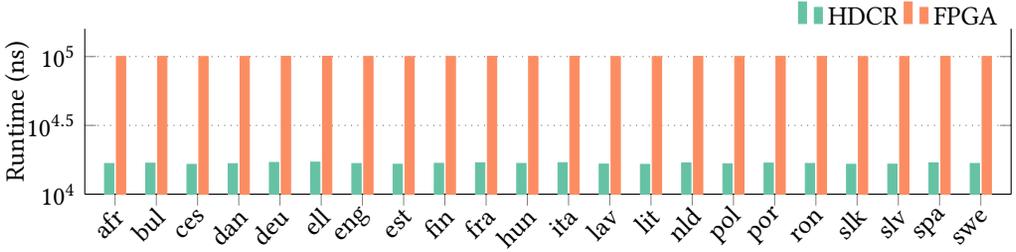
\begin{figure*}
\pgfplotsset{compat = newest}
\pgfplotsset{major grid style={dotted,aluminium2!50!black}}
\begin{tikzpicture}
\begin{axis}
[
    width=1\textwidth,
    ybar=1pt, 
    enlargelimits=0.035,
    enlarge y limits={upper, value=0.2},
    ylabel style={align=center},
    y label style={at={(-0.05,0.5)}},
    ylabel=Runtime (ns),
    legend style={draw=none, fill=none},
    bar width=3.5pt,
    legend columns=2,
    ymin=10000, 
    ymode = log,
    log basis y={10},
    height=0.27\textwidth,
    ymajorgrids=true,
    grid style=dashed,
    axis x line*=bottom,
    x tick label style={xshift=.4em,rotate=45,anchor=east},
    yminorticks=true,
    legend style={at={(1,1.2)},anchor=north east},
    xlabel={},
    every y tick scale label/.style={at={(yticklabel cs:1.2)}, anchor = north west, rotate = 0},
    symbolic x coords={afr, bul, ces, dan, deu, ell, eng, est, fin, fra, hun, ita, lav, lit, nld, pol, por, ron, slk, slv, spa, swe},
    xtick=data,
]


\addplot+ [blind_safe_one_scheme_three_colors] coordinates {
(afr,16701  ) 
(bul,16775 ) 
(ces,16454  )
(dan,16648 ) 
(deu,16999 ) 
(ell,17074 ) 
(eng,16684 ) 
(est,16479 ) 
(fin,16737 ) 
(fra,16894 ) 
(hun,16711 ) 
(ita,16934 ) 
(lav,16527 ) 
(lit,16454 ) 
(nld,16866) 
(pol,16634 ) 
(por,16803 )
(ron,16713) 
(slk,16473 ) 
(slv,16498) 
(spa,16877) 
(swe,16710 ) };  

\addplot+ [blind_safe_two_scheme_three_colors] coordinates {
(afr,100123.019428571)
(bul,100164.432)
(ces,99915.24)
(dan,100077.948)
(deu,100354.704)
(ell,100406.22)
(eng,100098.9)
(est,99947.292)
(fin,100150.272)
(fra,100270.044)
(hun,100126.62)
(ita,100300.884)
(lav,99978.72)
(lit,99915.78)
(nld,100242.24)
(pol,100057.968)
(por,100195.812)
(ron,100126.884)
(slk,99926.304)
(slv,99950.472)
(spa,100250.16)
(swe,100126.512)};


\legend{HDCR, FPGA}
.\end{axis}
\end{tikzpicture}
\caption{Runtime of the HDC inference on different platforms. The results are generated on average length input text for all languages.}
\label{fig:runtime_inference}
\end{figure*}

The runtime comparison for training and inference in HDCR and FPGA designs is presented in Fig.~\ref{fig:runtime_train} and Fig.~\ref{fig:runtime_inference}, respectively. The runtime, and also the energy consumption in the next section, for the training and inference phases are computed and reported separately because training is typically performed once and in advance. In contrast, the inference is performed more frequently in real-world applications. Therefore, the measured values for the inference should be regarded as having a higher relevance. Since the runtime depends on the number of letters in the input text, which varies for different languages, the evaluation is performed for each language.

On average (geomean), HDCR is an order of magnitude faster compared to the FPGA design. 
Note that the FPGA implementation we used for comparison is already optimized for a high degree of concurrency and parallelism. All hypervectors are stored in registers, and encoding an N-gram requires only a single clock cycle, \textit{i.e.}, all $N$ HVs are simultaneously permuted, and the \texttt{XOR} operation is performed directly in the same combinational path. This results in long combinational paths, which leads to a lower clock frequency of \SI{80}{\mega\hertz}. The massively parallel implementation of bit operations on the vectors also results in an enormous consumption of resources, limiting the given FPGA design to large devices, e.g., from the Virtex 7 series. Unlike the encoding operation, the similarity check module compares HVs sequentially and requires 8192 cycles to compare the query HV to a single class HV. This module is replicated 22 times to compare to all languages simultaneously.


In HDC training, only the encoding module is used to encode large training texts\footnote{The number of characters for the training texts was between 100000 and 200000.} into their respective class vectors. Despite the sequential rotation of hypervectors in HDCR, it outperforms the FPGA design by a geometric mean of $\approx10.2\times$ (see Fig.~\ref{fig:runtime_train}). This is mainly attributed to the smaller clock period in HDCR \SI{1}{\nano\second} compared to \SI{12.5}{\nano\second} in the FPGA design.

In HDC inference, due to the smaller input text\footnote{The number of characters for the test sentences was between 100 and 200 for all languages.}, the overall runtime of the FPGA design is largely dominated by the similarity checking module. We use an average sentence size per language generated from all 1000 test sentences per language in the test data set for this evaluation. Again, despite the sequentiality in population counting, HDCR on average (geomean) reduces the runtime by $\approx 6\times$ compared to the FPGA design (see Fig.~\ref{fig:runtime_inference}). This is because the FPGA design performs the vector comparison sequentially while HDCR compares in 512-bit chunks, in parallel across languages.


We also synthesized the hardware to an ASIC \SI{65}{\nano\meter} process using Cadence RTL Compiler to generate a performance comparison point consistent with the ASIC energy comparison point presented in the next section.  The best achievable clock speed was \SI{400}{\mega\hertz} which is approximately 5$\times$ faster than the FPGA implementation.   The silicon required an area of \SI{4.37}{\milli\meter}$^2$, which is quite substantial for a single function accelerator.  Given HDCR is more than $6\times$ faster for all operation modes and an ASIC implementation would be limited to the single task, HDCR provides a substantial benefit over custom silicon. 
For completeness of comparison, on the CPU machine, the training and inference modules require \SI{0.107e+03}{\sec} and \SI{0.095e+03}{\sec} for all 22 languages, which are four and seven orders of magnitude slower, respectively, compared to HDCR.
The reason for this significant performance gap is that CPU machine requires a huge amount of data shuffling between the memory and the core~\cite{thrifty}. Additionally, the HDCR performance gain is partially attributed to its custom in-memory compute units, \textit{e.g.,} the \texttt{XOR} implementation. Since the total energy consumption is dependent on the runtime, a similar trend is expected in the energy comparison of HDCR and the CPU machine. 

\subsection{Energy Consumption}
\label{subsec:energy}
In terms of energy consumption, HDCR is comparable to the FPGA design during the HDC training phase (see Fig.~\ref{fig:energy_train}) and $\approx 5.3 \times$ better during the inference phase. In the similarity checking module alone, HDCR reduces the energy consumption by $\approx 95 \times$ (see Fig.~\ref{fig:energy_inference}). However, this is masked by the roughly equivalent energy consumption of the encoder module in both designs. 
The dominant impact on the energy consumption for the HDCR encoding phase is attributed to the parallel implementation of the majority operation with RTM counters. This requires 8192 counters which enable the required number of parallel bit-write operations. Since the energy consumption for RTM is proportional to such write operations, it is correspondingly large for the encoding step.
The result presented in Fig.~\ref{fig:energy_train} shows the energy consumption during the training phase, which includes the encoder. While the results vary less than 1\% different between FPGA and HDCR, this analysis does not consider I/O energy associated with moving data to and from the accelerator.  
In both cases, the input letters need to be transferred from the main memory to the computing unit. While HDCR only needs the input letter to be read and sent to the RTM memory controller, the FPGA system must also forward the data on the bus to the FPGA implementation. 
This omission makes our results more conservative, but independent of how the external system interfaces the implementation. Regardless, the reduced inference-time energy allows the HDCR implementation to immediately realize a net energy benefit over the FPGA implementation as presented in Fig.~\ref{fig:energy_inference}. 

In the case of inference, the
similarity checking in HDRC requires a single counter per language, and the operation is performed only once. As soon as the bitwise comparison with the XOR operation is performed, the 1s in the resultant vector are aggregated using the TR operation and the RTM counter while the FPGA synthesizes a direct 1s counting circuit. 


To summarize, with regard to the overall energy efficiency, the HDCR implementation reduces the energy consumption by 5.3$\times$ (geomean). 


\begin{figure*}
\pgfplotsset{compat = newest}
\pgfplotsset{major grid style={dotted,aluminium2!50!black}}
\begin{tikzpicture}
\begin{axis}
[
    width=1\textwidth,
    ybar=1pt, 
    enlargelimits=0.035,
    enlarge y limits={upper, value=0.2},
    ylabel style={align=center},
    y label style={at={(-0.06,0.5)}},
    ylabel=Energy Consumption ($\mu$J),
    legend style={draw=none, fill=none},
    bar width=3.5pt,
    legend columns=2,
    ymin=30000, 
    ymode = log,
    log basis y={10},
    height=0.27\textwidth,
    ymajorgrids=true,
    grid style=dashed,
    axis x line*=bottom,
    x tick label style={xshift=.4em,rotate=45,anchor=east},
    yminorticks=true,
    legend style={at={(1,1.2)},anchor=north east},
    xlabel={},
    ytick scale label code/.code={\pgfmathparse{int(-#1)}$y \cdot 10^{\pgfmathresult}$},
    every y tick scale  every y tick scale label/.style={at={(yticklabel cs:1.2)}, anchor = north west, rotate = 0},
    symbolic x coords={afr, bul, ces, dan, deu, ell, eng, est, fin, fra, hun, ita, lav, lit, nld, pol, por, ron, slk, slv, spa, swe},
    xtick=data,
]

\addplot+ [blind_safe_one_scheme_three_colors] coordinates {
(afr,34903  ) 
(bul,39226 ) 
(ces, 35396  )
(dan,45958) 
(deu, 41968 ) 
(ell,47035) 
(eng,37574 ) 
(est,35346 ) 
(fin,37357 ) 
(fra,43788 ) 
(hun,42624 ) 
(ita,45405 ) 
(lav,38933 ) 
(lit,37941 ) 
(nld,35054  ) 
(pol,36421  ) 
(por,45907 )
(ron,44403 ) 
(slk,38118 ) 
(slv,41585 ) 
(spa,48924 ) 
(swe,34359  ) };  

\addplot+ [blind_safe_two_scheme_three_colors] coordinates {
(afr,35310.054  ) 
(bul,39977.91   ) 
(ces,36074.9232  )
(dan,46839.5244 ) 
(deu,42773.2668 ) 
(ell,47937.1104 ) 
(eng,38294.1264 ) 
(est,36023.4012  ) 
(fin,38073.3072 ) 
(fra,44627.5008 ) 
(hun,43441.1928 ) 
(ita,46275.7956 ) 
(lav,39679.9008 ) 
(lit,38668.842  ) 
(nld,35726.1732  ) 
(pol,37119.2388  ) 
(por,46787.556  )
(ron,45254.544  ) 
(slk,38849.3736 ) 
(slv,42382.4808 ) 
(spa,49862.508  ) 
(swe,35018.0712  ) };


\legend{HDCR, FPGA}
\end{axis}
\end{tikzpicture}
\caption{Energy consumption of the HDC training.}
\label{fig:energy_train}
\end{figure*}
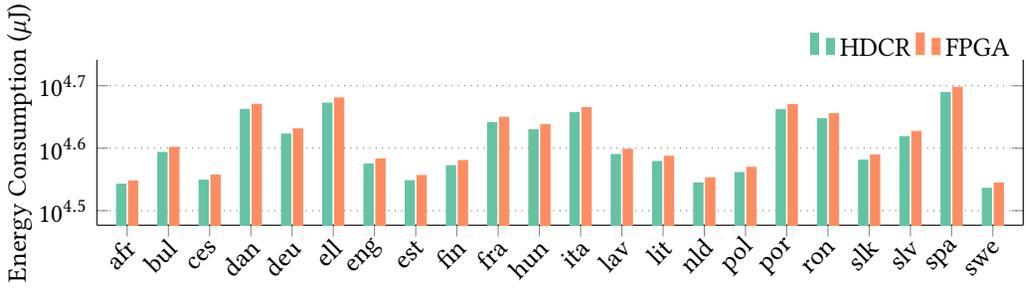

\begin{figure*}
\newcounter{groupcount}
\pgfplotsset{
    draw group line/.style n args={5}{
        after end axis/.append code={
            \setcounter{groupcount}{0}
            \pgfplotstableforeachcolumnelement{#1}\of\datatable\as\cell{%
                \def\temp{#2}
                \ifx\temp\cell
                    \ifnum\thegroupcount=0
                        \stepcounter{groupcount}
                        \pgfplotstablegetelem{\pgfplotstablerow}{X}\of\datatable
                        \coordinate [yshift=#4] (startgroup) at (axis cs:\pgfplotsretval,0);
                    \else
                        \pgfplotstablegetelem{\pgfplotstablerow}{X}\of\datatable
                        \coordinate [yshift=#4] (endgroup) at (axis cs:\pgfplotsretval,0);
                    \fi
                \else
                    \ifnum\thegroupcount=1
                        \setcounter{groupcount}{0}
                        \draw [
                            shorten >=-#5,
                            shorten <=-#5
                        ] (startgroup) -- node [anchor=base, yshift=0.5ex] {#3} (endgroup);
                    \fi
                \fi
            }
            \ifnum\thegroupcount=1
                        \setcounter{groupcount}{0}
                        \draw [
                            shorten >=-#5,
                            shorten <=-#5
                        ] (startgroup) -- node [anchor=base, yshift=0.5ex] {#3} (endgroup);
            \fi
        }
    }
}

\begin{tikzpicture}
\pgfplotstableread{
X	Name	Encoder	SimCheck	bench
1	HDCR	5.52  	0.242 	afr
2	FPGA	5.63    	25     	afr
4	HDCR	5.65	0.263 	bul
5	FPGA	5.767	25.2	bul
7	HDCR	4.89	0.276	ces
8	FPGA	4.99	25.2	ces
10	HDCR	5.36	0.276  	dan
11	FPGA	5.4992388	25	dan
13	HDCR	6.2	0.236	deu
14	FPGA	6.3571824	25	deu
16	HDCR	6.38	0.236	ell
17	FPGA	6.516882	25	ell
19	HDCR	5.43	0.263	eng
20	FPGA	5.56419	25	eng
22	HDCR	4.96	0.249	est
23	FPGA	5.0942052	25	est
25	HDCR	5.58	0.276	fin
26	FPGA	5.7234432	25	fin
28	HDCR	5.949	0.236	fra
29	FPGA	6.0947364	25	fra
31	HDCR	5.511	0.263	hun
32	FPGA	5.650122	25	hun
34	HDCR	6.05	0.276	ita
35	FPGA	6.1903404	25	ita
37	HDCR	5.07	0.263	lav
38	FPGA	5.191632	25	lav
40	HDCR	4.89	0.236	lit
41	FPGA	4.996518	25	lit
43	HDCR	5.87	0.249	nld
44	FPGA	6.008544	25	nld
46	HDCR	5.32	0.26	pol
47	FPGA	5.4373008	25	pol
49	HDCR	5.73	0.2631	por
50	FPGA	5.8646172	25	por
52	HDCR	5.511	0.249	ron
53	FPGA	5.6509404	25	ron
55	HDCR	4.927	0.29	slk
56	FPGA	5.0291424	25	slk
58	HDCR	5	0.25	slv
59	FPGA	5.1040632	25	slv
61	HDCR	5.913	0.29	spa
62	FPGA	6.033096	25	spa
64	HDCR	5.51	0.26	swe
65	FPGA	5.6497872	25	swe
}\datatable

\begin{axis}[
    width=1\textwidth,
    height=0.27\textwidth,
    ybar stacked,
    bar width=3.5pt,
    enlargelimits=0.020,
    enlarge y limits={upper, value=0.2},
    ylabel style={align=center},
    y label style={at={(-0.0,0.5)}},
    ylabel=Energy Consumption ($\mu$J),
    legend style={draw=none, fill=none},
    legend columns=2,
    legend style={at={(1,1.2)},anchor=north east},
    ymode=log,
    log basis y ={10}, 
    ymajorgrids=true,
    grid style=dashed,
    axis x line*=bottom,
    xtick=data,
    x tick label style={rotate=90,anchor=east, font=\tiny},
    yminorticks=true,
    xticklabels from table={\datatable}{Name},
    draw group line={bench}{afr}{afr}{-13ex}{3.6pt},
    draw group line={bench}{bul}{bul}{-13ex}{3.6pt},
    draw group line={bench}{ces}{ces}{-13ex}{3.6pt},
    draw group line={bench}{dan}{dan}{-13ex}{3.6pt},
    draw group line={bench}{deu}{deu}{-13ex}{3.6pt},
    draw group line={bench}{ell}{ell}{-13ex}{3.6pt},
    draw group line={bench}{eng}{eng}{-13ex}{3.6pt},
    draw group line={bench}{est}{est}{-13ex}{3.6pt},
    draw group line={bench}{fin}{fin}{-13ex}{3.6pt},
    draw group line={bench}{fra}{fra}{-13ex}{3.6pt},
    draw group line={bench}{hun}{hun}{-13ex}{3.6pt},
    draw group line={bench}{ita}{ita}{-13ex}{3.6pt},
    draw group line={bench}{lav}{lav}{-13ex}{3.6pt},
    draw group line={bench}{lit}{lit}{-13ex}{3.6pt},
    draw group line={bench}{nld}{nld}{-13ex}{3.6pt},
    draw group line={bench}{pol}{pol}{-13ex}{3.6pt},
    draw group line={bench}{por}{por}{-13ex}{3.6pt},
    draw group line={bench}{ron}{ron}{-13ex}{3.6pt},
    draw group line={bench}{slk}{slk}{-13ex}{3.6pt},
    draw group line={bench}{slv}{slv}{-13ex}{3.6pt},
    draw group line={bench}{spa}{spa}{-13ex}{3.6pt},
    draw group line={bench}{swe}{swe}{-13ex}{3.6pt},
    after end axis/.append code={
    \path [anchor=base east, yshift=0.5ex]
            (rel axis cs:0,0) node [yshift=-8.5ex] {Bench};
    }
]

\addplot+[blind_safe_one_scheme_three_colors] table [x=X, y=Encoder] {\datatable}; \addlegendentry{Encoder}
\addplot+[blind_safe_two_scheme_three_colors] table [x=X, y=SimCheck] {\datatable}; \addlegendentry{Sim\_Check}

\end{axis}
\end{tikzpicture}
\caption{Energy consumption of different modules in the HDC inference.}
\label{fig:energy_inference}
\end{figure*}

\subsection{Comparison between HDCR and PCM}

In the paper from Karunaratne et al.~\cite{inPCM_2020}, they propose to use the novel PCM memory to implement HDC. This work does not report the latency of their implementation, 
thus here we only show the energy comparison. Table~\ref{tab:hdcrvspcm} compares the inference energy consumption of the HDCR and PCM designs for an average-sized input text. Overall, HDCR outperforms the PCM design by 10.1$\times$ in the encoding module and 1.08$\times$ in the similarity search module. Although the PCM design reports dramatic reduction in the energy consumption in the similarity checking module, largely due to parallel multiplications and current accumulation in the crossbar architecture, its overall energy consumption is still higher than HDCR. This is due to the higher write energy of the memristive devices compared to RTM. Comparing with the \SI{65}{\nano\meter} CMOS-only design of the same reference, HDCR achieves a 51.6$\times$ improvement.

\begin{table}[tbh]
\centering
\caption{Average energy per query  }
\centering
\label{tab:hdcrvspcm}
  \begin{tabular}{c|c|c|c}
\toprule
 & Encoder & Sim\_Check & Total \\
\midrule
all-CMOS [\SI{}{\nano\joule}] & 1474 & 1110 & 2584\\
PCM [\SI{}{\nano\joule}] & 420.8 & 9.44 & 430.3\\
HDCR [\SI{}{\nano\joule}] & 41.4 & 8.67 & 50.07\\[1.5mm]
\hline 
Improvement (PCM / HDCR) & 10.1$\times$ & 1.08$\times$ & 8.59$\times$ \\ 
\bottomrule
\end{tabular}
\end{table}

\section{Related work}
\label{sec:related_work}
Hyperdimensional computing has been used for learning and classification problems in many application domains. Among others, HDC has been used for analogy-based reasoning~\cite{hdc_reasoning}, language classification~\cite{rahimi_TCS_17}, hand gesture and activity recognition~\cite{hdc_gesture}, text classification~\cite{hdc_voice}, and medical applications such as epileptic seizure detection~\cite{hdc_medical}. Although compared to conventional learning algorithms, HDC is considered lightweight, the dimensionality of the hypervectors still makes HDC resource-intensive, particularly on embedded and IoT devices. To improve the performance and energy consumption of the HDC frameworks, they have been accelerated on various platforms. These include: FPGAs~\cite{hdc_FPGA}, conventional CPUs and GPUs~\cite{hdc_cgpu}, and domain-specific accelerators~\cite{hdc_dsa, hdc_dsa2, hdc_dsa3}. Since HDC is a memory-intensive application and is based on simple mathematical and logical operations, the in-memory compute capabilities of emerging nonvolatile memory technologies can be exploited to accelerate it. 

Many recently proposed architectures conduct near- or in-memory computation using emerging nonvolatile memory technologies~\cite{eNVM}, typically tuned to leverage the strength of the particular memory technology and the intended application.  These works can be broadly categorized based on the underlying technology (phase-change memory (PCM), ReRAM, STT-MRAM), and further by how they conduct their processing (bitwise operations, arithmetic logic, vector multiplication).


Vector multiplication and arithmetic is a fundamental operation to many machine learning and neural network tasks.  In HDC, the same is applied in the encoding and similarity search modules to compute the n-gram hypervector and similarity score. Karunaratne et al.~\cite{inPCM_2020} implement dot-product operations using PCM in a crossbar architecture.  Using an on-chip network and DAC/ADC circuits, smaller multiply-accumulate subarrays are composed to realize larger dot-product results.  Other recent work conducts 8-bit multiply accumulate logic for convolutional neural networks by converting the values from digital to analog and uses analog crossbar computation to obtain the results~\cite{khaddam2022hermes}.  These, along with similar works leveraging ReRAM~\cite{inRRAM_2016, inRRAM_2018, hdcIM_2019} provide acceleration and improved energy consumption relative to GPU/CPU implementations, but offer limited flexibility for input size, limited accuracy associated with computation in the analog domain, and require additional area to interpret and accumulate the analog results. This makes such approaches unscalable for our target application.

Besides PCM and ReRAM, STT-MRAM technology can also be used for in-memory computation. For instance, HieIM~\cite{HieIM} and MLC-STT-CIM~\cite{MLC-STT-CIM} exploit customized STT-MRAM memories to conduct bitwise operations on memory contents and build arithmetic operations by combining bitwise operations. These designs offer energy and area benefits for simple large matrix operations such as convolution. Still, they are less efficient than other general PIM proposals and require customized cell designs, which are difficult to fabricate. A more efficient design in STT-CIM~\cite{STT-CiM} conducts computation by opening multiple rows and sensing the combined current on shared bitlines.  Using modified reference voltages at the sense amplifiers allows \texttt{OR}, \texttt{AND}, and \texttt{XOR} operations, which are then composed to realize arithmetic operations.  This is more efficient than prior designs since the additional circuitry is restricted to the sense amplifiers, and more realistic to fabricate since it does not modify the fundamental cell structures. Unfortunately, STT-MRAM designs require an access point and a fixed reference layer for every cell.  While some of this area's cost is mitigated by the use of crossbar architecture, the density is limited to the feature size of the access network. A similar density limitation exists for computation using other non-volatile memories~\cite{li2016pinatubo}, often with the added complication of limited endurance in the underlying memory cells.  In contrast, planar racetrack memories only need as many access points as the length of the DBC
, and in turn can achieve superior densities.

RTM was initially proposed as a secondary storage device~\cite{stuart1.0, stuart4.0}. However, due to its promising characteristics, particularly its best-case SRAM class latency and high energy efficiency, RTM has been considered for application at all levels --- from register file and instruction memory to SSDs --- in the memory stack. For instance, Mao and Wang \textit{et al.} have proposed RTM-based GPU register files to combat the high leakage and scalability problems of conventional SRAM-based register files~\cite{gpu_registerfile, gpu_rf}. Xu \textit{et al.} evaluated RTM at lower cache levels and reported an energy reduction of 69\% with comparable performance relative to an iso-capacity SRAM~\cite{fusedcache} and explored the impact of lightweight compression to allow independent shifting~\cite{7059042}. 
Venkatesan \textit{et al.} demonstrated RTM at last-level cache and reported significant improvements in area (6.4$\times$), energy (1.4$\times$) and performance (25\%)~\cite{tapecache}. Park advocates the usage of RTM instead of SSD for graph storage which not only expedites graph processing but also reduces 
energy by up-to 90\%~\cite{ssd}. Besides, RTMs have been proposed as scratchpad memories~\cite{lctes}, content addressable memories~\cite{content_addressable}, reconfigurable memories~\cite{reconfig_mem}, and even as network buffers~\cite{kline2018racetrack,7167194}. A recent review on RTMs covers more details on the latest developments in RTMs and provides an exhaustive list of references on the application of RTM in the memory subsystem~\cite{blaesing_2020}.

There are relatively fewer instances of processing-in-memory applied to racetrack memories.  The state-of-the-art offers three approaches: S-CNN~\cite{CNN_DWM}, DW-NN~\cite{DW-NN}, and PIRM~\cite{CORUSCANT}.  SPIM adds a dedicated processing unit utilizing skyrmions that can compute logical \texttt{OR} and \texttt{AND} operations.  Unfortunately, these operations require dedicated circuitry for a fixed number of operands, limiting the utility of the approach when more complex computation is required~\cite{CORUSCANT}. DW-NN uses dedicated racetrack pairs, which store data from either operand and compute logical functions by reading across the stacked magnetic domains.  Simple XOR operations are computed directly, and in concert with an additional precharge sensing amplifier, can be used to compute a \texttt{SUM} and \texttt{CARRY} for addition.  These results are then transferred to conventional racetracks, which can be shifted and summed to perform multiplication. Unfortunately, performance is bottlenecked in two places: first, the data must be read from the conventional racetracks to the paired racetracks one bit at a time.  Second, each bit position in the paired nanowires must be shifted under the access port, serializing the computation.  While the architecture offers an energy and throughput advantage compared to von Neumann, this serialization limits the utility of the approach.  Finally, PIRM offers a more generalized computation framework, utilizing a more capable PIM-enabled tile to compute arbitrary logical operations, addition, and multiplication.  PIRM accelerates computation by leveraging TR and multi-operands, and does not require specialized racetracks to do its work.  

Our cim-tile uses the same philosophy as PIRM, but is tuned for the operations needed to compute HDC.  Additionally, we explore new operations such as counters and majority determination. While prior work for HDC using in-memory PCM did not conduct a performance analysis, recent work for convolutional neural networks (CNNs) using a similar PIM approach did report 1.0 tera operations per second (TOPS) for 8-bit values.  This is a significant improvement over FPGA capabilities which can achieve 0.34 TOPS~\cite{jiang2019achieving}.  However, compute-in-racetrack memory approaches can outperform this PCM result by an order of magnitude, for example with S-CNN achieving 9.3 TOPS.  Given HDCR uses a similar mechanism to PIRM, and PIRM provides 26 TOPS for 8-bit CNN inference, we can expect similar order of magnitude speedups of HDCR over PCM for HDC applications. 

\section{Conclusions}
\label{sec:Conclusion}

The data dimensionality and mathematical properties of the HDC frameworks make them ideal fits for in-memory computations. Many conventional and emerging memory technologies allow (partial) implementation of the HDC framework in-memory. In this paper, we present a complete racetrack memory based HDC system, requiring near-negligible additional CMOS logic. Most of the HDC operations are implemented with the \emph{TR} operation that reports the number of 1s in the nanowire, exploiting its properties and magnetic domain (and domain wall) arrangements. For the majority and the population count operations, we propose RTM nanowires-based counters that are scaleable and area and energy-efficient compared to their CMOS counterparts. The hypervectors are organized in RTM in a way that allows maximum possible parallelism and minimum possible data movement. For the in-RTM computations, we dedicate one tile per subarray -- the cim-tile -- and make minimal but necessary changes to its peripheral circuitry. For the logic operations, a few additional multiplexing/selection gates are added to the row buffer circuitry to infer the transverse results into different HDC operations.
Our hardware customization and extensions are negligible compared to other memory technologies, \textit{e.g.}, the power-hungry ADC/DAC converters, etc., in memristive devices. For the language recognition use case, our proposed system, on average, consumes 5.33$\times$ and 8.59$\times$ less energy compared to the state-of-the-art FPGA and PCM-crossbar designs, respectively.

\section*{Acknowledgments}
This work was partially funded by the German Research Council (DFG) through the TraceSymm project (366764507) and the Co4RTM project (450944241), and by the NSF awards 1822085 and 2133267 and by the laboratory of physical sciences (LPS) and NSA.

\bibliographystyle{ACM-Reference-Format}
\balance
\bibliography{main.bib}


\begin{thebibliography}{70}


\ifx \showCODEN    \undefined \def \showCODEN     #1{\unskip}     \fi
\ifx \showDOI      \undefined \def \showDOI       #1{#1}\fi
\ifx \showISBNx    \undefined \def \showISBNx     #1{\unskip}     \fi
\ifx \showISBNxiii \undefined \def \showISBNxiii  #1{\unskip}     \fi
\ifx \showISSN     \undefined \def \showISSN      #1{\unskip}     \fi
\ifx \showLCCN     \undefined \def \showLCCN      #1{\unskip}     \fi
\ifx \shownote     \undefined \def \shownote      #1{#1}          \fi
\ifx \showarticletitle \undefined \def \showarticletitle #1{#1}   \fi
\ifx \showURL      \undefined \def \showURL       {\relax}        \fi
\providecommand\bibfield[2]{#2}
\providecommand\bibinfo[2]{#2}
\providecommand\natexlab[1]{#1}
\providecommand\showeprint[2][]{arXiv:#2}

\bibitem[\protect\citeauthoryear{Archer, Mappouras, Calderbank, and
  Sorin}{Archer et~al\mbox{.}}{2020}]%
        {archer2020foosball}
\bibfield{author}{\bibinfo{person}{Samantha Archer}, \bibinfo{person}{Georgios
  Mappouras}, \bibinfo{person}{Robert Calderbank}, {and}
  \bibinfo{person}{Daniel Sorin}.} \bibinfo{year}{2020}\natexlab{}.
\newblock \showarticletitle{Foosball coding: Correcting shift errors and bit
  flip errors in 3d racetrack memory}. In \bibinfo{booktitle}{\emph{2020 50th
  Annual IEEE/IFIP International Conference on Dependable Systems and Networks
  (DSN)}}. IEEE, \bibinfo{pages}{331--342}.
\newblock


\bibitem[\protect\citeauthoryear{Bl\"asing, Khan, Filippou, Garg, Hameed,
  Castrillon, and Parkin}{Bl\"asing et~al\mbox{.}}{2020}]%
        {blaesing_2020}
\bibfield{author}{\bibinfo{person}{Robin Bl\"asing}, \bibinfo{person}{Asif~Ali
  Khan}, \bibinfo{person}{Panagiotis~Ch. Filippou}, \bibinfo{person}{Chirag
  Garg}, \bibinfo{person}{Fazal Hameed}, \bibinfo{person}{Jeronimo Castrillon},
  {and} \bibinfo{person}{Stuart S.~P. Parkin}.}
  \bibinfo{year}{2020}\natexlab{}.
\newblock \showarticletitle{Magnetic Racetrack Memory: From Physics to the Cusp
  of Applications Within a Decade}.
\newblock \bibinfo{journal}{\emph{Proc. IEEE}} \bibinfo{volume}{108},
  \bibinfo{number}{8} (\bibinfo{year}{2020}), \bibinfo{pages}{1303--1321}.
\newblock
\urldef\tempurl%
\url{https://doi.org/10.1109/JPROC.2020.2975719}
\showDOI{\tempurl}


\bibitem[\protect\citeauthoryear{Bl{\"a}sing, Ma, Yang, Garg, Dejene,
  T~N’Diaye, Chen, Liu, and Parkin}{Bl{\"a}sing et~al\mbox{.}}{2018}]%
        {blasing2018exchange}
\bibfield{author}{\bibinfo{person}{Robin Bl{\"a}sing},
  \bibinfo{person}{Tianping Ma}, \bibinfo{person}{See-Hun Yang},
  \bibinfo{person}{Chirag Garg}, \bibinfo{person}{Fasil~Kidane Dejene},
  \bibinfo{person}{Alpha T~N’Diaye}, \bibinfo{person}{Gong Chen},
  \bibinfo{person}{Kai Liu}, {and} \bibinfo{person}{Stuart~SP Parkin}.}
  \bibinfo{year}{2018}\natexlab{}.
\newblock \showarticletitle{Exchange coupling torque in ferrimagnetic Co/Gd
  bilayer maximized near angular momentum compensation temperature}.
\newblock \bibinfo{journal}{\emph{Nature communications}} \bibinfo{volume}{9},
  \bibinfo{number}{1} (\bibinfo{year}{2018}), \bibinfo{pages}{1--8}.
\newblock


\bibitem[\protect\citeauthoryear{Burrello, Cavigelli, Schindler, Benini, and
  Rahimi}{Burrello et~al\mbox{.}}{2019}]%
        {hdc_medical}
\bibfield{author}{\bibinfo{person}{Alessio Burrello}, \bibinfo{person}{Lukas
  Cavigelli}, \bibinfo{person}{Kaspar Schindler}, \bibinfo{person}{Luca
  Benini}, {and} \bibinfo{person}{Abbas Rahimi}.}
  \bibinfo{year}{2019}\natexlab{}.
\newblock \showarticletitle{Laelaps: An Energy-Efficient Seizure Detection
  Algorithm from Long-term Human iEEG Recordings without False Alarms}. In
  \bibinfo{booktitle}{\emph{2019 Design, Automation Test in Europe Conference
  Exhibition (DATE)}}. \bibinfo{pages}{752--757}.
\newblock
\urldef\tempurl%
\url{https://doi.org/10.23919/DATE.2019.8715186}
\showDOI{\tempurl}


\bibitem[\protect\citeauthoryear{Datta, Antonio, Ison, and Rabaey}{Datta
  et~al\mbox{.}}{2019a}]%
        {hdcASIC_2019}
\bibfield{author}{\bibinfo{person}{Sohum Datta}, \bibinfo{person}{Ryan A.~G.
  Antonio}, \bibinfo{person}{Aldrin R.~S. Ison}, {and} \bibinfo{person}{Jan~M.
  Rabaey}.} \bibinfo{year}{2019}\natexlab{a}.
\newblock \showarticletitle{A Programmable Hyper-Dimensional Processor
  Architecture for Human-Centric IoT}.
\newblock \bibinfo{journal}{\emph{IEEE Journal on Emerging and Selected Topics
  in Circuits and Systems}} \bibinfo{volume}{9}, \bibinfo{number}{3}
  (\bibinfo{year}{2019}), \bibinfo{pages}{439--452}.
\newblock
\urldef\tempurl%
\url{https://doi.org/10.1109/JETCAS.2019.2935464}
\showDOI{\tempurl}


\bibitem[\protect\citeauthoryear{Datta, Antonio, Ison, and Rabaey}{Datta
  et~al\mbox{.}}{2019b}]%
        {hdc_cgpu}
\bibfield{author}{\bibinfo{person}{Sohum Datta}, \bibinfo{person}{Ryan A.~G.
  Antonio}, \bibinfo{person}{Aldrin R.~S. Ison}, {and} \bibinfo{person}{Jan~M.
  Rabaey}.} \bibinfo{year}{2019}\natexlab{b}.
\newblock \showarticletitle{A Programmable Hyper-Dimensional Processor
  Architecture for Human-Centric IoT}.
\newblock \bibinfo{journal}{\emph{IEEE Journal on Emerging and Selected Topics
  in Circuits and Systems}} \bibinfo{volume}{9}, \bibinfo{number}{3}
  (\bibinfo{year}{2019}), \bibinfo{pages}{439--452}.
\newblock
\urldef\tempurl%
\url{https://doi.org/10.1109/JETCAS.2019.2935464}
\showDOI{\tempurl}


\bibitem[\protect\citeauthoryear{Deng, Jiang, Zhang, Zhang, and Yang}{Deng
  et~al\mbox{.}}{2018}]%
        {deng2018dracc}
\bibfield{author}{\bibinfo{person}{Quan Deng}, \bibinfo{person}{Lei Jiang},
  \bibinfo{person}{Youtao Zhang}, \bibinfo{person}{Minxuan Zhang}, {and}
  \bibinfo{person}{Jun Yang}.} \bibinfo{year}{2018}\natexlab{}.
\newblock \showarticletitle{Dracc: a dram based accelerator for accurate cnn
  inference}. In \bibinfo{booktitle}{\emph{Proceedings of the 55th Annual
  Design Automation Conference}}. \bibinfo{pages}{1--6}.
\newblock


\bibitem[\protect\citeauthoryear{Ge and Parhi}{Ge and Parhi}{2020}]%
        {hdc_review}
\bibfield{author}{\bibinfo{person}{Lulu Ge} {and} \bibinfo{person}{Keshab~K
  Parhi}.} \bibinfo{year}{2020}\natexlab{}.
\newblock \showarticletitle{Classification using hyperdimensional computing: A
  review}.
\newblock \bibinfo{journal}{\emph{IEEE Circuits and Systems Magazine}}
  \bibinfo{volume}{20}, \bibinfo{number}{2} (\bibinfo{year}{2020}),
  \bibinfo{pages}{30--47}.
\newblock


\bibitem[\protect\citeauthoryear{Gupta, Morris, Imani, Ramkumar, Yu, Tiwari,
  Aksanli, and Rosing}{Gupta et~al\mbox{.}}{2020}]%
        {thrifty}
\bibfield{author}{\bibinfo{person}{Saransh Gupta}, \bibinfo{person}{Justin
  Morris}, \bibinfo{person}{Mohsen Imani}, \bibinfo{person}{Ranganathan
  Ramkumar}, \bibinfo{person}{Jeffrey Yu}, \bibinfo{person}{Aniket Tiwari},
  \bibinfo{person}{Baris Aksanli}, {and}
  \bibinfo{person}{Tajana~{\v{S}}imuni{\'c} Rosing}.}
  \bibinfo{year}{2020}\natexlab{}.
\newblock \showarticletitle{Thrifty: Training with hyperdimensional computing
  across flash hierarchy}. In \bibinfo{booktitle}{\emph{2020 IEEE/ACM
  International Conference On Computer Aided Design (ICCAD)}}. IEEE,
  \bibinfo{pages}{1--9}.
\newblock


\bibitem[\protect\citeauthoryear{Hassan, Halawani, Mohammad, and Saleh}{Hassan
  et~al\mbox{.}}{2021}]%
        {hdc_comp}
\bibfield{author}{\bibinfo{person}{Eman Hassan}, \bibinfo{person}{Yasmin
  Halawani}, \bibinfo{person}{Baker Mohammad}, {and} \bibinfo{person}{Hani
  Saleh}.} \bibinfo{year}{2021}\natexlab{}.
\newblock \showarticletitle{Hyper-Dimensional Computing Challenges and
  Opportunities for AI Applications}.
\newblock \bibinfo{journal}{\emph{IEEE Access}} (\bibinfo{year}{2021}).
\newblock


\bibitem[\protect\citeauthoryear{Hersche et~al\mbox{.}}{Hersche
  et~al\mbox{.}}{[n.d.]}]%
        {hdc_C}
\bibfield{author}{\bibinfo{person}{Michael Hersche} {et~al\mbox{.}}}
  \bibinfo{year}{[n.d.]}\natexlab{}.
\newblock \bibinfo{title}{HDlib}.
\newblock \bibinfo{howpublished}{\url{https://github.com/skurella/hdlib}}.
\newblock
\newblock
\shownote{Accessed: 2022-02-22.}


\bibitem[\protect\citeauthoryear{Hersche, del R.~Millán, Benini, and
  Rahimi}{Hersche et~al\mbox{.}}{2018}]%
        {2018svmvshdc}
\bibfield{author}{\bibinfo{person}{Michael Hersche}, \bibinfo{person}{José del
  R.~Millán}, \bibinfo{person}{Luca Benini}, {and} \bibinfo{person}{Abbas
  Rahimi}.} \bibinfo{year}{2018}\natexlab{}.
\newblock \bibinfo{title}{Exploring Embedding Methods in Binary
  Hyperdimensional Computing: A Case Study for Motor-Imagery based
  Brain-Computer Interfaces}.
\newblock
\newblock
\showeprint[arxiv]{1812.05705}~[eess.SP]


\bibitem[\protect\citeauthoryear{Ielmini and Wong}{Ielmini and Wong}{2018}]%
        {cim_resistive_2018}
\bibfield{author}{\bibinfo{person}{Daniele Ielmini} {and}
  \bibinfo{person}{H-S~Philip Wong}.} \bibinfo{year}{2018}\natexlab{}.
\newblock \showarticletitle{In-memory computing with resistive switching
  devices}.
\newblock \bibinfo{journal}{\emph{Nature Electronics}} \bibinfo{volume}{1},
  \bibinfo{number}{6} (\bibinfo{year}{2018}), \bibinfo{pages}{333--343}.
\newblock


\bibitem[\protect\citeauthoryear{Imani, Kong, Rahimi, and Rosing}{Imani
  et~al\mbox{.}}{2017}]%
        {hdc_voice}
\bibfield{author}{\bibinfo{person}{Mohsen Imani}, \bibinfo{person}{Deqian
  Kong}, \bibinfo{person}{Abbas Rahimi}, {and} \bibinfo{person}{Tajana
  Rosing}.} \bibinfo{year}{2017}\natexlab{}.
\newblock \showarticletitle{VoiceHD: Hyperdimensional Computing for Efficient
  Speech Recognition}. In \bibinfo{booktitle}{\emph{2017 IEEE International
  Conference on Rebooting Computing (ICRC)}}. \bibinfo{pages}{1--8}.
\newblock
\urldef\tempurl%
\url{https://doi.org/10.1109/ICRC.2017.8123650}
\showDOI{\tempurl}


\bibitem[\protect\citeauthoryear{Imani, Zou, Bosch, Rao, Salamat, Kumar, Kim,
  and Rosing}{Imani et~al\mbox{.}}{2021}]%
        {hdc_dsa3}
\bibfield{author}{\bibinfo{person}{Mohsen Imani}, \bibinfo{person}{Zhuowen
  Zou}, \bibinfo{person}{Samuel Bosch}, \bibinfo{person}{Sanjay~Anantha Rao},
  \bibinfo{person}{Sahand Salamat}, \bibinfo{person}{Venkatesh Kumar},
  \bibinfo{person}{Yeseong Kim}, {and} \bibinfo{person}{Tajana Rosing}.}
  \bibinfo{year}{2021}\natexlab{}.
\newblock \showarticletitle{Revisiting HyperDimensional Learning for FPGA and
  Low-Power Architectures}. In \bibinfo{booktitle}{\emph{2021 IEEE
  International Symposium on High-Performance Computer Architecture (HPCA)}}.
  \bibinfo{pages}{221--234}.
\newblock
\urldef\tempurl%
\url{https://doi.org/10.1109/HPCA51647.2021.00028}
\showDOI{\tempurl}


\bibitem[\protect\citeauthoryear{Jain, Ranjan, Roy, and Raghunathan}{Jain
  et~al\mbox{.}}{2018}]%
        {STT-CiM}
\bibfield{author}{\bibinfo{person}{Shubham Jain}, \bibinfo{person}{Ashish
  Ranjan}, \bibinfo{person}{Kaushik Roy}, {and} \bibinfo{person}{Anand
  Raghunathan}.} \bibinfo{year}{2018}\natexlab{}.
\newblock \showarticletitle{Computing in Memory With Spin-Transfer Torque
  Magnetic RAM}.
\newblock \bibinfo{journal}{\emph{IEEE Transactions on Very Large Scale
  Integration (VLSI) Systems}} \bibinfo{volume}{26}, \bibinfo{number}{3}
  (\bibinfo{year}{2018}), \bibinfo{pages}{470--483}.
\newblock
\urldef\tempurl%
\url{https://doi.org/10.1109/TVLSI.2017.2776954}
\showDOI{\tempurl}


\bibitem[\protect\citeauthoryear{Jiang, Sha, Zhang, Yang, Zhuge, Shi, and
  Hu}{Jiang et~al\mbox{.}}{2019}]%
        {jiang2019achieving}
\bibfield{author}{\bibinfo{person}{Weiwen Jiang}, \bibinfo{person}{Edwin H-M
  Sha}, \bibinfo{person}{Xinyi Zhang}, \bibinfo{person}{Lei Yang},
  \bibinfo{person}{Qingfeng Zhuge}, \bibinfo{person}{Yiyu Shi}, {and}
  \bibinfo{person}{Jingtong Hu}.} \bibinfo{year}{2019}\natexlab{}.
\newblock \showarticletitle{Achieving super-linear speedup across multi-fpga
  for real-time dnn inference}.
\newblock \bibinfo{journal}{\emph{ACM Transactions on Embedded Computing
  Systems (TECS)}} \bibinfo{volume}{18}, \bibinfo{number}{5s}
  (\bibinfo{year}{2019}), \bibinfo{pages}{67}.
\newblock


\bibitem[\protect\citeauthoryear{Kanerva}{Kanerva}{1996}]%
        {BSC}
\bibfield{author}{\bibinfo{person}{Pentti Kanerva}.}
  \bibinfo{year}{1996}\natexlab{}.
\newblock \showarticletitle{Binary spatter-coding of ordered K-tuples}. In
  \bibinfo{booktitle}{\emph{Artificial Neural Networks --- ICANN 96}}.
  \bibinfo{publisher}{Springer Berlin Heidelberg}, \bibinfo{address}{Berlin,
  Heidelberg}, \bibinfo{pages}{869--873}.
\newblock
\showISBNx{978-3-540-68684-2}


\bibitem[\protect\citeauthoryear{Kanerva}{Kanerva}{2009}]%
        {kanerva_2009}
\bibfield{author}{\bibinfo{person}{Pentti Kanerva}.}
  \bibinfo{year}{2009}\natexlab{}.
\newblock \showarticletitle{Hyperdimensional Computing: An Introduction to
  Computing in Distributed Representation with High-Dimensional Random
  Vectors}.
\newblock \bibinfo{journal}{\emph{Cognitive Computation}}  \bibinfo{volume}{1}
  (\bibinfo{date}{06} \bibinfo{year}{2009}), \bibinfo{pages}{139--159}.
\newblock
\urldef\tempurl%
\url{https://doi.org/10.1007/s12559-009-9009-8}
\showDOI{\tempurl}


\bibitem[\protect\citeauthoryear{Kanerva}{Kanerva}{2010}]%
        {hdc_reasoning}
\bibfield{author}{\bibinfo{person}{P. Kanerva}.}
  \bibinfo{year}{2010}\natexlab{}.
\newblock \showarticletitle{What We Mean When We Say "What's the Dollar of
  Mexico?": Prototypes and Mapping in Concept Space}. In
  \bibinfo{booktitle}{\emph{AAAI Fall Symposium: Quantum Informatics for
  Cognitive, Social, and Semantic Processes}}.
\newblock


\bibitem[\protect\citeauthoryear{Karunaratne, Gallo, Cherubini, Benini, Rahimi,
  and Sebastian}{Karunaratne et~al\mbox{.}}{2020}]%
        {inPCM_2020}
\bibfield{author}{\bibinfo{person}{Geethan Karunaratne},
  \bibinfo{person}{Manuel Gallo}, \bibinfo{person}{Giovanni Cherubini},
  \bibinfo{person}{Luca Benini}, \bibinfo{person}{Abbas Rahimi}, {and}
  \bibinfo{person}{Abu Sebastian}.} \bibinfo{year}{2020}\natexlab{}.
\newblock \showarticletitle{In-memory hyperdimensional computing}.
\newblock \bibinfo{journal}{\emph{Nature Electronics}}  \bibinfo{volume}{3}
  (\bibinfo{year}{2020}), \bibinfo{pages}{327--337}.
\newblock
\urldef\tempurl%
\url{https://doi.org/10.1038/s41928-020-0410-3}
\showDOI{\tempurl}


\bibitem[\protect\citeauthoryear{Khaddam-Aljameh, Stanisavljevic, Mas,
  Karunaratne, Br{\"a}ndli, Liu, Singh, M{\"u}ller, Egger, Petropoulos,
  et~al\mbox{.}}{Khaddam-Aljameh et~al\mbox{.}}{2022}]%
        {khaddam2022hermes}
\bibfield{author}{\bibinfo{person}{Riduan Khaddam-Aljameh},
  \bibinfo{person}{Milos Stanisavljevic}, \bibinfo{person}{Jordi~Fornt Mas},
  \bibinfo{person}{Geethan Karunaratne}, \bibinfo{person}{Matthias
  Br{\"a}ndli}, \bibinfo{person}{Feng Liu}, \bibinfo{person}{Abhairaj Singh},
  \bibinfo{person}{Silvia~M M{\"u}ller}, \bibinfo{person}{Urs Egger},
  \bibinfo{person}{Anastasios Petropoulos}, {et~al\mbox{.}}}
  \bibinfo{year}{2022}\natexlab{}.
\newblock \showarticletitle{HERMES-Core--A 1.59-TOPS/mm$^2$ PCM on 14-nm CMOS
  In-Memory Compute Core Using 300-ps/LSB Linearized CCO-Based ADCs}.
\newblock \bibinfo{journal}{\emph{IEEE Journal of Solid-State Circuits}}
  (\bibinfo{year}{2022}).
\newblock


\bibitem[\protect\citeauthoryear{Khan, Hameed, Bl{\"a}sing, Parkin, and
  Castrillon}{Khan et~al\mbox{.}}{2019a}]%
        {shiftsreduce}
\bibfield{author}{\bibinfo{person}{Asif~Ali Khan}, \bibinfo{person}{Fazal
  Hameed}, \bibinfo{person}{Robin Bl{\"a}sing}, \bibinfo{person}{Stuart~SP
  Parkin}, {and} \bibinfo{person}{Jeronimo Castrillon}.}
  \bibinfo{year}{2019}\natexlab{a}.
\newblock \showarticletitle{Shiftsreduce: Minimizing shifts in racetrack memory
  4.0}.
\newblock \bibinfo{journal}{\emph{ACM Transactions on Architecture and Code
  Optimization (TACO)}} \bibinfo{volume}{16}, \bibinfo{number}{4}
  (\bibinfo{year}{2019}), \bibinfo{pages}{1--23}.
\newblock


\bibitem[\protect\citeauthoryear{Khan, Rink, Hameed, and Castrillon}{Khan
  et~al\mbox{.}}{2019b}]%
        {lctes}
\bibfield{author}{\bibinfo{person}{Asif~Ali Khan}, \bibinfo{person}{Norman~A.
  Rink}, \bibinfo{person}{Fazal Hameed}, {and} \bibinfo{person}{Jeronimo
  Castrillon}.} \bibinfo{year}{2019}\natexlab{b}.
\newblock \showarticletitle{Optimizing Tensor Contractions for Embedded Devices
  with Racetrack Memory Scratch-Pads}. In \bibinfo{booktitle}{\emph{Proceedings
  of the 20th ACM SIGPLAN/SIGBED International Conference on Languages,
  Compilers, and Tools for Embedded Systems}} (Phoenix, AZ, USA)
  \emph{(\bibinfo{series}{LCTES 2019})}. \bibinfo{publisher}{Association for
  Computing Machinery}, \bibinfo{address}{New York, NY, USA},
  \bibinfo{pages}{5–18}.
\newblock
\showISBNx{9781450367240}
\urldef\tempurl%
\url{https://doi.org/10.1145/3316482.3326351}
\showDOI{\tempurl}


\bibitem[\protect\citeauthoryear{Kim, Kim, Hirata, Oh, Tono, Kim, Okuno, Ham,
  Kim, Go, et~al\mbox{.}}{Kim et~al\mbox{.}}{2017}]%
        {dwspeed}
\bibfield{author}{\bibinfo{person}{Kab-Jin Kim}, \bibinfo{person}{Se~Kwon Kim},
  \bibinfo{person}{Yuushou Hirata}, \bibinfo{person}{Se-Hyeok Oh},
  \bibinfo{person}{Takayuki Tono}, \bibinfo{person}{Duck-Ho Kim},
  \bibinfo{person}{Takaya Okuno}, \bibinfo{person}{Woo~Seung Ham},
  \bibinfo{person}{Sanghoon Kim}, \bibinfo{person}{Gyoungchoon Go},
  {et~al\mbox{.}}} \bibinfo{year}{2017}\natexlab{}.
\newblock \showarticletitle{Fast domain wall motion in the vicinity of the
  angular momentum compensation temperature of ferrimagnets}.
\newblock \bibinfo{journal}{\emph{Nature materials}} \bibinfo{volume}{16},
  \bibinfo{number}{12} (\bibinfo{year}{2017}), \bibinfo{pages}{1187--1192}.
\newblock


\bibitem[\protect\citeauthoryear{Kim, Imani, Moshiri, and Rosing}{Kim
  et~al\mbox{.}}{2020}]%
        {hdc_dsa}
\bibfield{author}{\bibinfo{person}{Yeseong Kim}, \bibinfo{person}{Mohsen
  Imani}, \bibinfo{person}{Niema Moshiri}, {and} \bibinfo{person}{Tajana
  Rosing}.} \bibinfo{year}{2020}\natexlab{}.
\newblock \showarticletitle{Geniehd: Efficient dna pattern matching accelerator
  using hyperdimensional computing}. In \bibinfo{booktitle}{\emph{2020 Design,
  Automation \& Test in Europe Conference \& Exhibition (DATE)}}. IEEE,
  \bibinfo{pages}{115--120}.
\newblock


\bibitem[\protect\citeauthoryear{Kline, Xu, Melhem, and Jones}{Kline
  et~al\mbox{.}}{2015}]%
        {7167194}
\bibfield{author}{\bibinfo{person}{Donald Kline}, \bibinfo{person}{Haifeng Xu},
  \bibinfo{person}{Rami Melhem}, {and} \bibinfo{person}{Alex~K. Jones}.}
  \bibinfo{year}{2015}\natexlab{}.
\newblock \showarticletitle{Domain-wall memory buffer for low-energy NoCs}. In
  \bibinfo{booktitle}{\emph{2015 52nd ACM/EDAC/IEEE Design Automation
  Conference (DAC)}}. \bibinfo{pages}{1--6}.
\newblock
\urldef\tempurl%
\url{https://doi.org/10.1145/2744769.2744826}
\showDOI{\tempurl}


\bibitem[\protect\citeauthoryear{Kline, Xu, Melhem, and Jones}{Kline
  et~al\mbox{.}}{2018}]%
        {kline2018racetrack}
\bibfield{author}{\bibinfo{person}{Donald Kline}, \bibinfo{person}{Haifeng Xu},
  \bibinfo{person}{Rami Melhem}, {and} \bibinfo{person}{Alex~K Jones}.}
  \bibinfo{year}{2018}\natexlab{}.
\newblock \showarticletitle{Racetrack Queues for Extremely Low-Energy FIFOs}.
\newblock \bibinfo{journal}{\emph{IEEE Transactions on Very Large Scale
  Integration (VLSI) Systems}} \bibinfo{number}{99} (\bibinfo{year}{2018}),
  \bibinfo{pages}{1--14}.
\newblock


\bibitem[\protect\citeauthoryear{Koehn}{Koehn}{2005}]%
        {LR_inference_corpus}
\bibfield{author}{\bibinfo{person}{Philipp Koehn}.}
  \bibinfo{year}{2005}\natexlab{}.
\newblock \showarticletitle{Europarl: A Parallel Corpus for Statistical Machine
  Translation}.
\newblock


\bibitem[\protect\citeauthoryear{Li, Yan, and Li}{Li et~al\mbox{.}}{2019}]%
        {eNVM}
\bibfield{author}{\bibinfo{person}{Bing Li}, \bibinfo{person}{Bonan Yan}, {and}
  \bibinfo{person}{Hai Li}.} \bibinfo{year}{2019}\natexlab{}.
\newblock \showarticletitle{An Overview of In-Memory Processing with Emerging
  Non-Volatile Memory for Data-Intensive Applications}. In
  \bibinfo{booktitle}{\emph{Proceedings of the 2019 on Great Lakes Symposium on
  VLSI}} (Tysons Corner, VA, USA) \emph{(\bibinfo{series}{GLSVLSI '19})}.
  \bibinfo{publisher}{Association for Computing Machinery},
  \bibinfo{address}{New York, NY, USA}, \bibinfo{pages}{381–386}.
\newblock
\showISBNx{9781450362528}
\urldef\tempurl%
\url{https://doi.org/10.1145/3299874.3319452}
\showDOI{\tempurl}


\bibitem[\protect\citeauthoryear{Li, Wu, Rahimi, Li, Rusch, Lin, Hsu, Sabry,
  Eryilmaz, Sohn, Chiu, Chen, Wu, Shieh, Yeh, Rabaey, Mitra, and Wong}{Li
  et~al\mbox{.}}{2016a}]%
        {inRRAM_2016}
\bibfield{author}{\bibinfo{person}{Haitong Li}, \bibinfo{person}{Tony~F. Wu},
  \bibinfo{person}{Abbas Rahimi}, \bibinfo{person}{Kai-Shin Li},
  \bibinfo{person}{Miles Rusch}, \bibinfo{person}{Chang-Hsien Lin},
  \bibinfo{person}{Juo-Luen Hsu}, \bibinfo{person}{Mohamed~M. Sabry},
  \bibinfo{person}{S.~Burc Eryilmaz}, \bibinfo{person}{Joon Sohn},
  \bibinfo{person}{Wen-Cheng Chiu}, \bibinfo{person}{Min-Cheng Chen},
  \bibinfo{person}{Tsung-Ta Wu}, \bibinfo{person}{Jia-Min Shieh},
  \bibinfo{person}{Wen-Kuan Yeh}, \bibinfo{person}{Jan~M. Rabaey},
  \bibinfo{person}{Subhasish Mitra}, {and} \bibinfo{person}{H.-S.~Philip
  Wong}.} \bibinfo{year}{2016}\natexlab{a}.
\newblock \showarticletitle{Hyperdimensional computing with 3D VRRAM in-memory
  kernels: Device-architecture co-design for energy-efficient, error-resilient
  language recognition}. In \bibinfo{booktitle}{\emph{2016 IEEE International
  Electron Devices Meeting (IEDM)}}. \bibinfo{pages}{16.1.1--16.1.4}.
\newblock
\urldef\tempurl%
\url{https://doi.org/10.1109/IEDM.2016.7838428}
\showDOI{\tempurl}


\bibitem[\protect\citeauthoryear{Li, Xu, Zou, Zhao, Lu, and Xie}{Li
  et~al\mbox{.}}{2016b}]%
        {li2016pinatubo}
\bibfield{author}{\bibinfo{person}{Shuangchen Li}, \bibinfo{person}{Cong Xu},
  \bibinfo{person}{Qiaosha Zou}, \bibinfo{person}{Jishen Zhao},
  \bibinfo{person}{Yu Lu}, {and} \bibinfo{person}{Yuan Xie}.}
  \bibinfo{year}{2016}\natexlab{b}.
\newblock \showarticletitle{Pinatubo: A processing-in-memory architecture for
  bulk bitwise operations in emerging non-volatile memories}. In
  \bibinfo{booktitle}{\emph{Proceedings of the 53rd Annual Design Automation
  Conference}}. \bibinfo{pages}{1--6}.
\newblock


\bibitem[\protect\citeauthoryear{Liu, Gu, Chen, Kang, Hu, Zhuge, and Sha}{Liu
  et~al\mbox{.}}{2017}]%
        {CNN_DWM}
\bibfield{author}{\bibinfo{person}{Bicheng Liu}, \bibinfo{person}{Shouzhen Gu},
  \bibinfo{person}{Mingsong Chen}, \bibinfo{person}{Wang Kang},
  \bibinfo{person}{Jingtong Hu}, \bibinfo{person}{Qingfeng Zhuge}, {and}
  \bibinfo{person}{Edwin H-M Sha}.} \bibinfo{year}{2017}\natexlab{}.
\newblock \showarticletitle{An efficient racetrack memory-based
  Processing-in-memory architecture for convolutional neural networks}. In
  \bibinfo{booktitle}{\emph{2017 IEEE International Symposium on Parallel and
  Distributed Processing with Applications and 2017 IEEE International
  Conference on Ubiquitous Computing and Communications (ISPA/IUCC)}}. IEEE,
  \bibinfo{pages}{383--390}.
\newblock


\bibitem[\protect\citeauthoryear{Liu, Ma, Zhu, Wang, and Yang}{Liu
  et~al\mbox{.}}{2019}]%
        {hdcIM_2019}
\bibfield{author}{\bibinfo{person}{Jialong Liu}, \bibinfo{person}{Mingyuan Ma},
  \bibinfo{person}{Zhenhua Zhu}, \bibinfo{person}{Yu Wang}, {and}
  \bibinfo{person}{Huazhong Yang}.} \bibinfo{year}{2019}\natexlab{}.
\newblock \showarticletitle{HDC-IM: Hyperdimensional Computing In-Memory
  Architecture based on RRAM}. In \bibinfo{booktitle}{\emph{2019 26th IEEE
  International Conference on Electronics, Circuits and Systems (ICECS)}}.
  \bibinfo{pages}{450--453}.
\newblock
\urldef\tempurl%
\url{https://doi.org/10.1109/ICECS46596.2019.8964906}
\showDOI{\tempurl}


\bibitem[\protect\citeauthoryear{Mao, Wen, Zhang, Chen, and Li}{Mao
  et~al\mbox{.}}{2017}]%
        {gpu_rf}
\bibfield{author}{\bibinfo{person}{Mengjie Mao}, \bibinfo{person}{Wujie Wen},
  \bibinfo{person}{Yaojun Zhang}, \bibinfo{person}{Yiran Chen}, {and}
  \bibinfo{person}{Hai Li}.} \bibinfo{year}{2017}\natexlab{}.
\newblock \showarticletitle{An Energy-Efficient GPGPU Register File
  Architecture Using Racetrack Memory}.
\newblock \bibinfo{journal}{\emph{IEEE Trans. Comput.}} \bibinfo{volume}{66},
  \bibinfo{number}{9} (\bibinfo{year}{2017}), \bibinfo{pages}{1478--1490}.
\newblock


\bibitem[\protect\citeauthoryear{Montagna, Rahimi, Benatti, Rossi, and
  Benini}{Montagna et~al\mbox{.}}{2018}]%
        {hdc_dsa2}
\bibfield{author}{\bibinfo{person}{Fabio Montagna}, \bibinfo{person}{Abbas
  Rahimi}, \bibinfo{person}{Simone Benatti}, \bibinfo{person}{Davide Rossi},
  {and} \bibinfo{person}{Luca Benini}.} \bibinfo{year}{2018}\natexlab{}.
\newblock \showarticletitle{PULP-HD: Accelerating Brain-Inspired
  High-Dimensional Computing on a Parallel Ultra-Low Power Platform}. In
  \bibinfo{booktitle}{\emph{2018 55th ACM/ESDA/IEEE Design Automation
  Conference (DAC)}}. \bibinfo{pages}{1--6}.
\newblock
\urldef\tempurl%
\url{https://doi.org/10.1109/DAC.2018.8465801}
\showDOI{\tempurl}


\bibitem[\protect\citeauthoryear{Morris, Imani, Bosch, Thomas, Shu, and
  Rosing}{Morris et~al\mbox{.}}{2019}]%
        {hdc_gesture}
\bibfield{author}{\bibinfo{person}{Justin Morris}, \bibinfo{person}{Mohsen
  Imani}, \bibinfo{person}{Samuel Bosch}, \bibinfo{person}{Anthony Thomas},
  \bibinfo{person}{Helen Shu}, {and} \bibinfo{person}{Tajana Rosing}.}
  \bibinfo{year}{2019}\natexlab{}.
\newblock \showarticletitle{CompHD: Efficient Hyperdimensional Computing Using
  Model Compression}. In \bibinfo{booktitle}{\emph{2019 IEEE/ACM International
  Symposium on Low Power Electronics and Design (ISLPED)}}.
  \bibinfo{pages}{1--6}.
\newblock
\urldef\tempurl%
\url{https://doi.org/10.1109/ISLPED.2019.8824908}
\showDOI{\tempurl}


\bibitem[\protect\citeauthoryear{Ollivier, Kline~Jr., Kawsher, Melhem, Banja,
  and Jones}{Ollivier et~al\mbox{.}}{2019}]%
        {ollivier2019dsn}
\bibfield{author}{\bibinfo{person}{Sébastien Ollivier},
  \bibinfo{person}{Donald Kline~Jr.}, \bibinfo{person}{Roxy Kawsher},
  \bibinfo{person}{Rami Melhem}, \bibinfo{person}{Sanjukta Banja}, {and}
  \bibinfo{person}{Alex~K. Jones}.} \bibinfo{year}{2019}\natexlab{}.
\newblock \showarticletitle{Leveraging Transverse Reads to Correct Alignment
  Faults in Domain Wall Memories}. In \bibinfo{booktitle}{\emph{Proceedings of
  the IEEE/IFIP Dependable Systems and Networks Conference (DSN)}}.
  \bibinfo{address}{Portland, OR}.
\newblock


\bibitem[\protect\citeauthoryear{Ollivier, Longofono, Dutta, Hu, Bhanja, and
  Jones}{Ollivier et~al\mbox{.}}{2021}]%
        {CORUSCANT}
\bibfield{author}{\bibinfo{person}{Sébastien Ollivier},
  \bibinfo{person}{Stephen Longofono}, \bibinfo{person}{Prayash Dutta},
  \bibinfo{person}{Jingtong Hu}, \bibinfo{person}{Sanjukta Bhanja}, {and}
  \bibinfo{person}{Alex~K. Jones}.} \bibinfo{year}{2021}\natexlab{}.
\newblock \showarticletitle{PIRM: Processing In Racetrack Memories}.
\newblock \bibinfo{journal}{\emph{arXiV}} (\bibinfo{date}{August}
  \bibinfo{year}{2021}).
\newblock
\showeprint{2108.00000}


\bibitem[\protect\citeauthoryear{Pan, Ouyang, Zhao, Kang, Yin, Zhang, Zhao, and
  Wei}{Pan et~al\mbox{.}}{2018}]%
        {MLC-STT-CIM}
\bibfield{author}{\bibinfo{person}{Yu Pan}, \bibinfo{person}{Peng Ouyang},
  \bibinfo{person}{Yinglin Zhao}, \bibinfo{person}{Wang Kang},
  \bibinfo{person}{Shouyi Yin}, \bibinfo{person}{Youguang Zhang},
  \bibinfo{person}{Weisheng Zhao}, {and} \bibinfo{person}{Shaojun Wei}.}
  \bibinfo{year}{2018}\natexlab{}.
\newblock \showarticletitle{A Multilevel Cell STT-MRAM-Based Computing
  In-Memory Accelerator for Binary Convolutional Neural Network}.
\newblock \bibinfo{journal}{\emph{IEEE Transactions on Magnetics}}
  \bibinfo{volume}{54}, \bibinfo{number}{11} (\bibinfo{year}{2018}),
  \bibinfo{pages}{1--5}.
\newblock
\urldef\tempurl%
\url{https://doi.org/10.1109/TMAG.2018.2848625}
\showDOI{\tempurl}


\bibitem[\protect\citeauthoryear{Park, Yoo, Lee, and Li}{Park
  et~al\mbox{.}}{2014}]%
        {ssd}
\bibfield{author}{\bibinfo{person}{E. Park}, \bibinfo{person}{S. Yoo},
  \bibinfo{person}{S. Lee}, {and} \bibinfo{person}{H. Li}.}
  \bibinfo{year}{2014}\natexlab{}.
\newblock \showarticletitle{Accelerating graph computation with racetrack
  memory and pointer-assisted graph representation}. In
  \bibinfo{booktitle}{\emph{2014 Design, Automation Test in Europe Conference
  Exhibition (DATE)}}. \bibinfo{pages}{1--4}.
\newblock
\showISSN{1530-1591}
\urldef\tempurl%
\url{https://doi.org/10.7873/DATE.2014.172}
\showDOI{\tempurl}


\bibitem[\protect\citeauthoryear{Parkin, Hayashi, and Thomas}{Parkin
  et~al\mbox{.}}{2008}]%
        {stuart1.0}
\bibfield{author}{\bibinfo{person}{S. Parkin}, \bibinfo{person}{M. Hayashi},
  {and} \bibinfo{person}{L. Thomas}.} \bibinfo{year}{2008}\natexlab{}.
\newblock \showarticletitle{{Magnetic Domain-Wall Racetrack Memory}}.
\newblock   \bibinfo{volume}{320} (\bibinfo{date}{05} \bibinfo{year}{2008}),
  \bibinfo{pages}{190--194}.
\newblock


\bibitem[\protect\citeauthoryear{Parkin and Yang}{Parkin and Yang}{2015}]%
        {stuart4.0}
\bibfield{author}{\bibinfo{person}{Stuart Parkin} {and}
  \bibinfo{person}{See-Hun Yang}.} \bibinfo{year}{2015}\natexlab{}.
\newblock \showarticletitle{{Memory on the Racetrack}}.
\newblock   \bibinfo{volume}{10} (\bibinfo{date}{03} \bibinfo{year}{2015}),
  \bibinfo{pages}{195--198}.
\newblock


\bibitem[\protect\citeauthoryear{Parveen, He, Angizi, and Fan}{Parveen
  et~al\mbox{.}}{2018}]%
        {HieIM}
\bibfield{author}{\bibinfo{person}{Farhana Parveen}, \bibinfo{person}{Zhezhi
  He}, \bibinfo{person}{Shaahin Angizi}, {and} \bibinfo{person}{Deliang Fan}.}
  \bibinfo{year}{2018}\natexlab{}.
\newblock \showarticletitle{HieIM: Highly Flexible in-Memory Computing Using
  STT MRAM}. In \bibinfo{booktitle}{\emph{Proceedings of the 23rd Asia and
  South Pacific Design Automation Conference}} (Jeju, Republic of Korea)
  \emph{(\bibinfo{series}{ASPDAC '18})}. \bibinfo{publisher}{IEEE Press},
  \bibinfo{pages}{361–366}.
\newblock


\bibitem[\protect\citeauthoryear{Puebla, Kim, Kondou, and Otani}{Puebla
  et~al\mbox{.}}{2020}]%
        {spindevices}
\bibfield{author}{\bibinfo{person}{Jorge Puebla}, \bibinfo{person}{Junyeon
  Kim}, \bibinfo{person}{Kouta Kondou}, {and} \bibinfo{person}{Yoshichika
  Otani}.} \bibinfo{year}{2020}\natexlab{}.
\newblock \showarticletitle{Spintronic devices for energy-efficient data
  storage and energy harvesting}.
\newblock \bibinfo{journal}{\emph{Communications Materials}}
  \bibinfo{volume}{1}, \bibinfo{number}{1} (\bibinfo{year}{2020}),
  \bibinfo{pages}{1--9}.
\newblock


\bibitem[\protect\citeauthoryear{Quasthoff, Richter, and Biemann}{Quasthoff
  et~al\mbox{.}}{2006}]%
        {LR_training_corpus}
\bibfield{author}{\bibinfo{person}{Uwe Quasthoff}, \bibinfo{person}{Matthias
  Richter}, {and} \bibinfo{person}{Chris Biemann}.}
  \bibinfo{year}{2006}\natexlab{}.
\newblock \showarticletitle{Corpus Portal for Search in Monolingual Corpora}.
\newblock \bibinfo{journal}{\emph{Proceedings of LREC-06}} (\bibinfo{date}{01}
  \bibinfo{year}{2006}).
\newblock


\bibitem[\protect\citeauthoryear{Rahimi et~al\mbox{.}}{Rahimi
  et~al\mbox{.}}{[n.d.]}]%
        {git_sources}
\bibfield{author}{\bibinfo{person}{Abbas Rahimi} {et~al\mbox{.}}}
  \bibinfo{year}{[n.d.]}\natexlab{}.
\newblock \bibinfo{title}{HDC Language Recognition}.
\newblock
  \bibinfo{howpublished}{\url{https://github.com/abbas-rahimi/HDC-Language-Recognition}}.
\newblock
\newblock
\shownote{Accessed: 2021-07-05.}


\bibitem[\protect\citeauthoryear{Rahimi, Datta, Kleyko, Frady, Olshausen,
  Kanerva, and Rabaey}{Rahimi et~al\mbox{.}}{2017}]%
        {rahimi_TCS_17}
\bibfield{author}{\bibinfo{person}{Abbas Rahimi}, \bibinfo{person}{Sohum
  Datta}, \bibinfo{person}{Denis Kleyko}, \bibinfo{person}{Edward~Paxon Frady},
  \bibinfo{person}{Bruno Olshausen}, \bibinfo{person}{Pentti Kanerva}, {and}
  \bibinfo{person}{Jan~M. Rabaey}.} \bibinfo{year}{2017}\natexlab{}.
\newblock \showarticletitle{High-Dimensional Computing as a Nanoscalable
  Paradigm}.
\newblock \bibinfo{journal}{\emph{IEEE Transactions on Circuits and Systems I:
  Regular Papers}} \bibinfo{volume}{64}, \bibinfo{number}{9}
  (\bibinfo{year}{2017}), \bibinfo{pages}{2508--2521}.
\newblock
\urldef\tempurl%
\url{https://doi.org/10.1109/TCSI.2017.2705051}
\showDOI{\tempurl}


\bibitem[\protect\citeauthoryear{Rahimi, Kanerva, and Rabaey}{Rahimi
  et~al\mbox{.}}{2016}]%
        {rahimi_ISLPED_16}
\bibfield{author}{\bibinfo{person}{Abbas Rahimi}, \bibinfo{person}{Pentti
  Kanerva}, {and} \bibinfo{person}{Jan~M. Rabaey}.}
  \bibinfo{year}{2016}\natexlab{}.
\newblock \showarticletitle{A Robust and Energy-Efficient Classifier Using
  Brain-Inspired Hyperdimensional Computing}. In
  \bibinfo{booktitle}{\emph{Proceedings of the 2016 International Symposium on
  Low Power Electronics and Design}} (San Francisco Airport, CA, USA)
  \emph{(\bibinfo{series}{ISLPED '16})}. \bibinfo{publisher}{Association for
  Computing Machinery}, \bibinfo{address}{New York, NY, USA},
  \bibinfo{pages}{64–69}.
\newblock
\showISBNx{9781450341851}
\urldef\tempurl%
\url{https://doi.org/10.1145/2934583.2934624}
\showDOI{\tempurl}


\bibitem[\protect\citeauthoryear{Riente, Turvani, Vacca, and Graziano}{Riente
  et~al\mbox{.}}{2021}]%
        {cim_rtm_2021}
\bibfield{author}{\bibinfo{person}{Fabrizio Riente}, \bibinfo{person}{Giovanna
  Turvani}, \bibinfo{person}{Marco Vacca}, {and} \bibinfo{person}{Mariagrazia
  Graziano}.} \bibinfo{year}{2021}\natexlab{}.
\newblock \showarticletitle{Parallel Computation in the Racetrack Memory}.
\newblock \bibinfo{journal}{\emph{IEEE Transactions on Emerging Topics in
  Computing}} (\bibinfo{year}{2021}), \bibinfo{pages}{1--1}.
\newblock
\urldef\tempurl%
\url{https://doi.org/10.1109/TETC.2021.3078061}
\showDOI{\tempurl}


\bibitem[\protect\citeauthoryear{Roxy, Ollivier, Hoque, Longofono, Jones, and
  Bhanja}{Roxy et~al\mbox{.}}{2020}]%
        {roxy2020novel}
\bibfield{author}{\bibinfo{person}{Kawsher Roxy},
  \bibinfo{person}{S{\'e}bastien Ollivier}, \bibinfo{person}{Arifa Hoque},
  \bibinfo{person}{Stephen Longofono}, \bibinfo{person}{Alex~K Jones}, {and}
  \bibinfo{person}{Sanjukta Bhanja}.} \bibinfo{year}{2020}\natexlab{}.
\newblock \showarticletitle{A Novel Transverse Read Technique for Domain-Wall
  “Racetrack” Memories}.
\newblock \bibinfo{journal}{\emph{IEEE Transactions on Nanotechnology}}
  \bibinfo{volume}{19} (\bibinfo{year}{2020}), \bibinfo{pages}{648--652}.
\newblock


\bibitem[\protect\citeauthoryear{Salamat, Imani, Khaleghi, and Rosing}{Salamat
  et~al\mbox{.}}{2019}]%
        {hdc_FPGA}
\bibfield{author}{\bibinfo{person}{Sahand Salamat}, \bibinfo{person}{Mohsen
  Imani}, \bibinfo{person}{Behnam Khaleghi}, {and} \bibinfo{person}{Tajana
  Rosing}.} \bibinfo{year}{2019}\natexlab{}.
\newblock \showarticletitle{F5-HD: Fast Flexible FPGA-Based Framework for
  Refreshing Hyperdimensional Computing} \emph{(\bibinfo{series}{FPGA '19})}.
  \bibinfo{publisher}{Association for Computing Machinery},
  \bibinfo{address}{New York, NY, USA}, \bibinfo{pages}{53–62}.
\newblock
\showISBNx{9781450361378}
\urldef\tempurl%
\url{https://doi.org/10.1145/3289602.3293913}
\showDOI{\tempurl}


\bibitem[\protect\citeauthoryear{Salamat, Imani, and Rosing}{Salamat
  et~al\mbox{.}}{2020}]%
        {HDCFPGA}
\bibfield{author}{\bibinfo{person}{Sahand Salamat}, \bibinfo{person}{Mohsen
  Imani}, {and} \bibinfo{person}{Tajana Rosing}.}
  \bibinfo{year}{2020}\natexlab{}.
\newblock \showarticletitle{Accelerating hyperdimensional computing on fpgas by
  exploiting computational reuse}.
\newblock \bibinfo{journal}{\emph{IEEE Trans. Comput.}} \bibinfo{volume}{69},
  \bibinfo{number}{8} (\bibinfo{year}{2020}), \bibinfo{pages}{1159--1171}.
\newblock


\bibitem[\protect\citeauthoryear{Schlegel, Neubert, and Protzel}{Schlegel
  et~al\mbox{.}}{2020}]%
        {vsas}
\bibfield{author}{\bibinfo{person}{Kenny Schlegel}, \bibinfo{person}{Peer
  Neubert}, {and} \bibinfo{person}{Peter Protzel}.}
  \bibinfo{year}{2020}\natexlab{}.
\newblock \showarticletitle{A comparison of vector symbolic architectures}.
\newblock \bibinfo{journal}{\emph{arXiv preprint arXiv:2001.11797}}
  (\bibinfo{year}{2020}).
\newblock


\bibitem[\protect\citeauthoryear{Shafiee, Nag, Muralimanohar, Balasubramonian,
  Strachan, Hu, Williams, and Srikumar}{Shafiee et~al\mbox{.}}{2016}]%
        {isaac_2016}
\bibfield{author}{\bibinfo{person}{Ali Shafiee}, \bibinfo{person}{Anirban Nag},
  \bibinfo{person}{Naveen Muralimanohar}, \bibinfo{person}{Rajeev
  Balasubramonian}, \bibinfo{person}{John~Paul Strachan}, \bibinfo{person}{Miao
  Hu}, \bibinfo{person}{R.~Stanley Williams}, {and} \bibinfo{person}{Vivek
  Srikumar}.} \bibinfo{year}{2016}\natexlab{}.
\newblock \showarticletitle{ISAAC: A Convolutional Neural Network Accelerator
  with In-Situ Analog Arithmetic in Crossbars}. In
  \bibinfo{booktitle}{\emph{2016 ACM/IEEE 43rd Annual International Symposium
  on Computer Architecture (ISCA)}}. \bibinfo{pages}{14--26}.
\newblock
\urldef\tempurl%
\url{https://doi.org/10.1109/ISCA.2016.12}
\showDOI{\tempurl}


\bibitem[\protect\citeauthoryear{Strubell, Ganesh, and McCallum}{Strubell
  et~al\mbox{.}}{2019}]%
        {NLP_models_energy}
\bibfield{author}{\bibinfo{person}{Emma Strubell}, \bibinfo{person}{Ananya
  Ganesh}, {and} \bibinfo{person}{Andrew McCallum}.}
  \bibinfo{year}{2019}\natexlab{}.
\newblock \bibinfo{title}{Energy and Policy Considerations for Deep Learning in
  NLP}.
\newblock
\newblock
\showeprint[arxiv]{1906.02243}~[cs.CL]


\bibitem[\protect\citeauthoryear{Thompson, Greenewald, Lee, and Manso}{Thompson
  et~al\mbox{.}}{2020}]%
        {ML_compute_resources}
\bibfield{author}{\bibinfo{person}{Neil~C. Thompson}, \bibinfo{person}{Kristjan
  Greenewald}, \bibinfo{person}{Keeheon Lee}, {and} \bibinfo{person}{Gabriel~F.
  Manso}.} \bibinfo{year}{2020}\natexlab{}.
\newblock \bibinfo{title}{The Computational Limits of Deep Learning}.
\newblock
\newblock
\showeprint[arxiv]{2007.05558}~[cs.LG]


\bibitem[\protect\citeauthoryear{Vatanen, V{\"a}yrynen, and Virpioja}{Vatanen
  et~al\mbox{.}}{2010}]%
        {2010langmodel}
\bibfield{author}{\bibinfo{person}{Tommi Vatanen}, \bibinfo{person}{Jaakko~J
  V{\"a}yrynen}, {and} \bibinfo{person}{Sami Virpioja}.}
  \bibinfo{year}{2010}\natexlab{}.
\newblock \showarticletitle{Language Identification of Short Text Segments with
  N-gram Models.}. In \bibinfo{booktitle}{\emph{LREC}}. Citeseer.
\newblock


\bibitem[\protect\citeauthoryear{Venkatesan, Kozhikkottu, Augustine,
  Raychowdhury, Roy, and Raghunathan}{Venkatesan et~al\mbox{.}}{2012}]%
        {tapecache}
\bibfield{author}{\bibinfo{person}{Rangharajan Venkatesan},
  \bibinfo{person}{Vivek Kozhikkottu}, \bibinfo{person}{Charles Augustine},
  \bibinfo{person}{Arijit Raychowdhury}, \bibinfo{person}{Kaushik Roy}, {and}
  \bibinfo{person}{Anand Raghunathan}.} \bibinfo{year}{2012}\natexlab{}.
\newblock \showarticletitle{TapeCache: A High Density, Energy Efficient Cache
  Based on Domain Wall Memory} \emph{(\bibinfo{series}{ISLPED '12})}.
  \bibinfo{publisher}{ACM}, \bibinfo{address}{New York, NY, USA},
  \bibinfo{pages}{185--190}.
\newblock
\showISBNx{978-1-4503-1249-3}
\urldef\tempurl%
\url{https://doi.org/10.1145/2333660.2333707}
\showDOI{\tempurl}


\bibitem[\protect\citeauthoryear{Venkatesan, Ramasubramanian, Venkataramani,
  Roy, and Raghunathan}{Venkatesan et~al\mbox{.}}{2014}]%
        {stag}
\bibfield{author}{\bibinfo{person}{Rangharajan Venkatesan},
  \bibinfo{person}{Shankar~Ganesh Ramasubramanian}, \bibinfo{person}{Swagath
  Venkataramani}, \bibinfo{person}{Kaushik Roy}, {and} \bibinfo{person}{Anand
  Raghunathan}.} \bibinfo{year}{2014}\natexlab{}.
\newblock \showarticletitle{STAG: Spintronic-Tape Architecture for GPGPU Cache
  Hierarchies}. In \bibinfo{booktitle}{\emph{Proceeding of the 41st Annual
  International Symposium on Computer Architecuture}} (Minneapolis, Minnesota,
  USA) \emph{(\bibinfo{series}{ISCA ’14})}. \bibinfo{publisher}{IEEE Press},
  \bibinfo{pages}{253–264}.
\newblock
\showISBNx{9781479943944}


\bibitem[\protect\citeauthoryear{Venkatesan, Sharad, Roy, and
  Raghunathan}{Venkatesan et~al\mbox{.}}{2013}]%
        {DWM_Tapestri}
\bibfield{author}{\bibinfo{person}{Rangharajan Venkatesan},
  \bibinfo{person}{Mrigank Sharad}, \bibinfo{person}{Kaushik Roy}, {and}
  \bibinfo{person}{Anand Raghunathan}.} \bibinfo{year}{2013}\natexlab{}.
\newblock \showarticletitle{DWM-TAPESTRI-an energy efficient all-spin cache
  using domain wall shift based writes}. In \bibinfo{booktitle}{\emph{Proc. of
  DATE}}. \bibinfo{pages}{1825--1830}.
\newblock


\bibitem[\protect\citeauthoryear{Wang, Liang, Zhang, Xie, Sun, Liu, Wang, and
  Li}{Wang et~al\mbox{.}}{2016}]%
        {gpu_registerfile}
\bibfield{author}{\bibinfo{person}{Shuo Wang}, \bibinfo{person}{Yun Liang},
  \bibinfo{person}{Chao Zhang}, \bibinfo{person}{Xiaolong Xie},
  \bibinfo{person}{Guangyu Sun}, \bibinfo{person}{Yongpan Liu},
  \bibinfo{person}{Yu Wang}, {and} \bibinfo{person}{Xiuhong Li}.}
  \bibinfo{year}{2016}\natexlab{}.
\newblock \showarticletitle{Performance-centric register file design for GPUs
  using racetrack memory}. In \bibinfo{booktitle}{\emph{2016 21st Asia and
  South Pacific Design Automation Conference (ASP-DAC)}}.
  \bibinfo{pages}{25--30}.
\newblock
\showISSN{2153-697X}
\urldef\tempurl%
\url{https://doi.org/10.1109/ASPDAC.2016.7427984}
\showDOI{\tempurl}


\bibitem[\protect\citeauthoryear{Wu, Li, Huang, Rahimi, Hills, Hodson, Hwang,
  Rabaey, Wong, Shulaker, and Mitra}{Wu et~al\mbox{.}}{2018}]%
        {inRRAM_2018}
\bibfield{author}{\bibinfo{person}{Tony~F. Wu}, \bibinfo{person}{Haitong Li},
  \bibinfo{person}{Ping-Chen Huang}, \bibinfo{person}{Abbas Rahimi},
  \bibinfo{person}{Gage Hills}, \bibinfo{person}{Bryce Hodson},
  \bibinfo{person}{William Hwang}, \bibinfo{person}{Jan~M. Rabaey},
  \bibinfo{person}{H.-S.~Philip Wong}, \bibinfo{person}{Max~M. Shulaker}, {and}
  \bibinfo{person}{Subhasish Mitra}.} \bibinfo{year}{2018}\natexlab{}.
\newblock \showarticletitle{Hyperdimensional Computing Exploiting Carbon
  Nanotube FETs, Resistive RAM, and Their Monolithic 3D Integration}.
\newblock \bibinfo{journal}{\emph{IEEE Journal of Solid-State Circuits}}
  \bibinfo{volume}{53}, \bibinfo{number}{11} (\bibinfo{year}{2018}),
  \bibinfo{pages}{3183--3196}.
\newblock
\urldef\tempurl%
\url{https://doi.org/10.1109/JSSC.2018.2870560}
\showDOI{\tempurl}


\bibitem[\protect\citeauthoryear{Xu, Alkabani, Melhem, and Jones}{Xu
  et~al\mbox{.}}{2016}]%
        {fusedcache}
\bibfield{author}{\bibinfo{person}{H. Xu}, \bibinfo{person}{Y. Alkabani},
  \bibinfo{person}{R. Melhem}, {and} \bibinfo{person}{A.~K. Jones}.}
  \bibinfo{year}{2016}\natexlab{}.
\newblock \showarticletitle{FusedCache: A Naturally Inclusive, Racetrack
  Memory, Dual-Level Private Cache}.
\newblock \bibinfo{journal}{\emph{IEEE Transactions on Multi-Scale Computing
  Systems}} \bibinfo{volume}{2}, \bibinfo{number}{2} (\bibinfo{date}{April}
  \bibinfo{year}{2016}), \bibinfo{pages}{69--82}.
\newblock
\showISSN{2332-7766}
\urldef\tempurl%
\url{https://doi.org/10.1109/TMSCS.2016.2536020}
\showDOI{\tempurl}


\bibitem[\protect\citeauthoryear{Xu, Li, Melhem, and Jones}{Xu
  et~al\mbox{.}}{2015}]%
        {7059042}
\bibfield{author}{\bibinfo{person}{Haifeng Xu}, \bibinfo{person}{Yong Li},
  \bibinfo{person}{Rami Melhem}, {and} \bibinfo{person}{Alex~K. Jones}.}
  \bibinfo{year}{2015}\natexlab{}.
\newblock \showarticletitle{Multilane Racetrack caches: Improving efficiency
  through compression and independent shifting}. In
  \bibinfo{booktitle}{\emph{The 20th Asia and South Pacific Design Automation
  Conference}}. \bibinfo{pages}{417--422}.
\newblock
\urldef\tempurl%
\url{https://doi.org/10.1109/ASPDAC.2015.7059042}
\showDOI{\tempurl}


\bibitem[\protect\citeauthoryear{Yu, Wang, Chen, Fei, Weng, Zhao, and Wei}{Yu
  et~al\mbox{.}}{2014}]%
        {DW-NN}
\bibfield{author}{\bibinfo{person}{Hao Yu}, \bibinfo{person}{Yuhao Wang},
  \bibinfo{person}{Shuai Chen}, \bibinfo{person}{Wei Fei},
  \bibinfo{person}{Chuliang Weng}, \bibinfo{person}{Junfeng Zhao}, {and}
  \bibinfo{person}{Zhulin Wei}.} \bibinfo{year}{2014}\natexlab{}.
\newblock \showarticletitle{Energy efficient in-memory machine learning for
  data intensive image-processing by non-volatile domain-wall memory}. In
  \bibinfo{booktitle}{\emph{2014 19th Asia and South Pacific Design Automation
  Conference (ASP-DAC)}}. \bibinfo{pages}{191--196}.
\newblock
\urldef\tempurl%
\url{https://doi.org/10.1109/ASPDAC.2014.6742888}
\showDOI{\tempurl}


\bibitem[\protect\citeauthoryear{Zhang, Sun, Zhang, Zhang, Zhao, Wang, Liang,
  Liu, Wang, and Shu}{Zhang et~al\mbox{.}}{2015}]%
        {hifi}
\bibfield{author}{\bibinfo{person}{Chao Zhang}, \bibinfo{person}{Guangyu Sun},
  \bibinfo{person}{Xian Zhang}, \bibinfo{person}{Weiqi Zhang},
  \bibinfo{person}{Weisheng Zhao}, \bibinfo{person}{Tao Wang},
  \bibinfo{person}{Yun Liang}, \bibinfo{person}{Yongpan Liu},
  \bibinfo{person}{Yu Wang}, {and} \bibinfo{person}{Jiwu Shu}.}
  \bibinfo{year}{2015}\natexlab{}.
\newblock \showarticletitle{Hi-fi playback: Tolerating position errors in shift
  operations of racetrack memory}. In \bibinfo{booktitle}{\emph{ACM SIGARCH
  Computer Architecture News}}, Vol.~\bibinfo{volume}{43-3}. ACM,
  \bibinfo{pages}{694--706}.
\newblock


\bibitem[\protect\citeauthoryear{Zhang, Zhao, Klein, Ravelsona, and
  Chappert}{Zhang et~al\mbox{.}}{2012a}]%
        {content_addressable}
\bibfield{author}{\bibinfo{person}{Y. Zhang}, \bibinfo{person}{W. Zhao},
  \bibinfo{person}{J. Klein}, \bibinfo{person}{D. Ravelsona}, {and}
  \bibinfo{person}{C. Chappert}.} \bibinfo{year}{2012}\natexlab{a}.
\newblock \showarticletitle{Ultra-High Density Content Addressable Memory Based
  on Current Induced Domain Wall Motion in Magnetic Track}.
\newblock \bibinfo{journal}{\emph{IEEE Transactions on Magnetics}}
  \bibinfo{volume}{48}, \bibinfo{number}{11} (\bibinfo{date}{Nov}
  \bibinfo{year}{2012}), \bibinfo{pages}{3219--3222}.
\newblock
\showISSN{0018-9464}
\urldef\tempurl%
\url{https://doi.org/10.1109/TMAG.2012.2198876}
\showDOI{\tempurl}


\bibitem[\protect\citeauthoryear{Zhang, Zhao, Ravelosona, Klein, Kim, and
  Chappert}{Zhang et~al\mbox{.}}{2012b}]%
        {zhang2012perpendicular}
\bibfield{author}{\bibinfo{person}{Yue Zhang}, \bibinfo{person}{WS Zhao},
  \bibinfo{person}{Dafin{\'e} Ravelosona}, \bibinfo{person}{J-O Klein},
  \bibinfo{person}{Joo-Von Kim}, {and} \bibinfo{person}{Claude Chappert}.}
  \bibinfo{year}{2012}\natexlab{b}.
\newblock \showarticletitle{Perpendicular-magnetic-anisotropy CoFeB racetrack
  memory}.
\newblock \bibinfo{journal}{\emph{Journal of Applied Physics}}
  \bibinfo{volume}{111}, \bibinfo{number}{9} (\bibinfo{year}{2012}),
  \bibinfo{pages}{093925}.
\newblock


\bibitem[\protect\citeauthoryear{Zhao, Romdhane, Zhang, Klein, and
  Ravelosona}{Zhao et~al\mbox{.}}{2013}]%
        {reconfig_mem}
\bibfield{author}{\bibinfo{person}{W. Zhao}, \bibinfo{person}{N.~Ben Romdhane},
  \bibinfo{person}{Y. Zhang}, \bibinfo{person}{J. Klein}, {and}
  \bibinfo{person}{D. Ravelosona}.} \bibinfo{year}{2013}\natexlab{}.
\newblock \showarticletitle{Racetrack memory based reconfigurable computing}.
  In \bibinfo{booktitle}{\emph{2013 IEEE Faible Tension Faible Consommation}}.
  \bibinfo{pages}{1--4}.
\newblock
\urldef\tempurl%
\url{https://doi.org/10.1109/FTFC.2013.6577771}
\showDOI{\tempurl}


\end{thebibliography}

\newpage
\thispagestyle{empty}
\section*{Acronyms}
\begin{acronym}
\acro{AP}[AP]{\emph{access port}}
\acro{AM}[AM]{\emph{associative memory}}
\acro{CIM}[CIM]{\emph{compute in memory}}
\acro{DW}[DW]{\emph{domain wall}}
\acro{DBC}[DBC]{\emph{domain wall block cluster}}
\acro{DWM}[DWM]{\emph{domain wall memory}}
\acro{HDC}[HDC]{\emph{hyperdimensional computing}}
\acro{HDCR}[HDCR]{\emph{HyperDimensional Computing in Racetrack}}
\acro{HV}[HV]{\emph{hypervector}}
\acro{IM}[IM]{\emph{item memory}}
\acro{LR}[LR]{\emph{language recognition}}
\acro{MTJ}[MTJ]{\emph{magnetic tunnel junction}}
\acro{PCM}[PCM]{\emph{phase change memory}}
\acro{PG}[PG]{\emph{processing group}}
\acro{RTM}[RTM]{\emph{racetrack memory}}
\acro{TR}[TR]{\emph{transverse read}}
\acro{TRD}[TRD]{\emph{transverse read distance}}
\acro{TW}[TW]{\emph{transverse write}}
\end{acronym}

\end{document}